\documentclass[lettersize,journal]{IEEEtran}
\usepackage{amsmath,amsfonts}
\usepackage{algorithmic}
\usepackage{algorithm}
\usepackage{array}
\usepackage[caption=false,font=normalsize,labelfont=sf,textfont=sf]{subfig}
\usepackage{textcomp}
\usepackage{stfloats}
\usepackage{url}
\usepackage{verbatim}
\usepackage{graphicx}
\usepackage{cite}
\usepackage[accsupp]{axessibility} 
\usepackage[bottom]{footmisc}
\usepackage{booktabs}
\usepackage{multirow}
\usepackage{graphicx}
\usepackage{xcolor}
\usepackage{colortbl}

\definecolor{Gray}{gray}{0.94}
\definecolor{liGray}{gray}{0.5}

\usepackage[normalem]{ulem}
\useunder{\uline}{\ul}{}
\usepackage{xspace}
\newcommand{\method}{\texttt{Animate-X++}\xspace}
\newcommand{\benchmark}{\texttt{$A^2$Bench}\xspace}

\newcommand{\tocite}[1]{\textcolor{red}{[TO CITE]}}

\newlength\savewidth\newcommand\shline{\noalign{\global\savewidth\arrayrulewidth
  \global\arrayrulewidth 1pt}\hline\noalign{\global\arrayrulewidth\savewidth}}

\begin{document}

\title{Animate-X++: Universal Character Image Animation with Dynamic Backgrounds}


\author{Shuai Tan, Biao Gong, Zhuoxin Liu, Yan Wang, Xi Chen, Yifan Feng, Hengshuang Zhao\IEEEauthorrefmark{2}

\IEEEcompsocitemizethanks{
\IEEEcompsocthanksitem{\IEEEauthorrefmark{2}Corresponding author.}
\IEEEcompsocthanksitem Shuai Tan, Xi Chen and Hengshuang Zhao are with School of Computing and Data Science, The University of Hong Kong, Hong Kong. E-mail: tanshuai@hku.hk, xichen.csai@gmail.com, hszhao@cs.hku.hk.
\IEEEcompsocthanksitem Biao Gong is with Ant Group, China. E-mail: a.biao.gong@gmail.com.
\IEEEcompsocthanksitem Zhuoxin Liu is with College of Letters and Science, The University of Wisconsin-Madison, United States. E-mail: zliu2464@wisc.edu.
\IEEEcompsocthanksitem Yan Wang is with College of Computer Science, University of North Carolina at Chapel Hill, United States. E-mail: wangyanscu@hotmail.com.
\IEEEcompsocthanksitem Yifan Feng is with School of Software, Tsinghua University, China. E-mail: evanfeng97@gmail.com.
}
\thanks{Preliminary version of this work was published in ICLR 2025\cite{animatex}.}
}

\markboth{Journal of \LaTeX\ Class Files,~Vol.~14, No.~8, August~2021}%
{Shell \MakeLowercase{\textit{et al.}}: A Sample Article Using IEEEtran.cls for IEEE Journals}


\maketitle

\begin{abstract}
Character image animation, which generates high-quality videos from a reference image and target pose sequence, has seen significant progress in recent years. However, most existing methods only apply to human figures, which usually do not generalize well on anthropomorphic characters commonly used in industries like gaming and entertainment. Furthermore, previous methods could only generate videos with static backgrounds, which limits the realism of the videos. For the first challenge, our in-depth analysis suggests to attribute this limitation to their insufficient modeling of motion, which is unable to comprehend the movement pattern of the driving video, thus imposing a pose sequence rigidly onto the target character. To this end, this paper proposes \method, a universal animation framework based on DiT for various character types (collectively named \texttt{X}), including anthropomorphic characters. To enhance motion representation, we introduce the Pose Indicator, which captures comprehensive motion pattern from the driving video through both implicit and explicit manner. The former leverages CLIP visual features of a driving video to extract its gist of motion, like the overall movement pattern and temporal relations among motions, while the latter strengthens the generalization of DiT by simulating possible inputs in advance that may arise during inference. For the second challenge, we introduce a multi-task training strategy that jointly trains the animation and Text-Image-to-Video (TI2V) tasks. Combined with the proposed partial parameter training, this approach achieves not only character animation but also text-driven background dynamics, making the videos more realistic. Moreover, we introduce a new Animated Anthropomorphic Benchmark (\benchmark) to evaluate the performance of \method on universal and widely applicable animation images. Extensive experiments demonstrate the superiority and effectiveness of \method. Our project page is available at: \url{https://lucaria-academy.github.io/Animate-X++/}.
\end{abstract}


\section{Introduction}
\label{sec: intro}
\IEEEPARstart{C}{haracter} image animation~\cite{yang2018pose,zablotskaia2019dwnet} is a compelling and challenging task that aims to generate lifelike, high-quality videos from a reference image and a target pose sequence. A modern image animation method shall ideally not only \textit{balance} the identity preservation and motion consistency, but also animate the surrounding background to create a more immersive and realistic visual experience, which contribute to the promise of broad utilization~\cite{Animateanyone,magicanimate,magicdance,jiang2022text2human}.
The phenomenal successes of GAN~\cite{goodfellow2014generative,yu2023bidirectionally,zhang2022exploring} and generative diffusion models~\cite{imagenvideo,DDPM,guo2023animatediff} have reshaped the performance of character animation generation. 
Nevertheless, most existing methods only apply to the human-specific character domain, and they are confined to generating videos with static backgrounds, which severely limits the realism of the output. In practice, the concept of \textit{“character”} encompasses a much broader concept than human, including anthropomorphic figures in cartoons and games, collectively referred to as \texttt{X}, which are more desirable in gaming, film, short videos, etc.
The difficulty in extending current models to these domains can be attributed to two main factors: {(1)} the predominantly human-centered nature of available datasets with static backgrounds, and {(2)} the limited generalization capabilities of current motion representations.

\begin{figure*}[t]
  \centering
\includegraphics[width=1\linewidth]{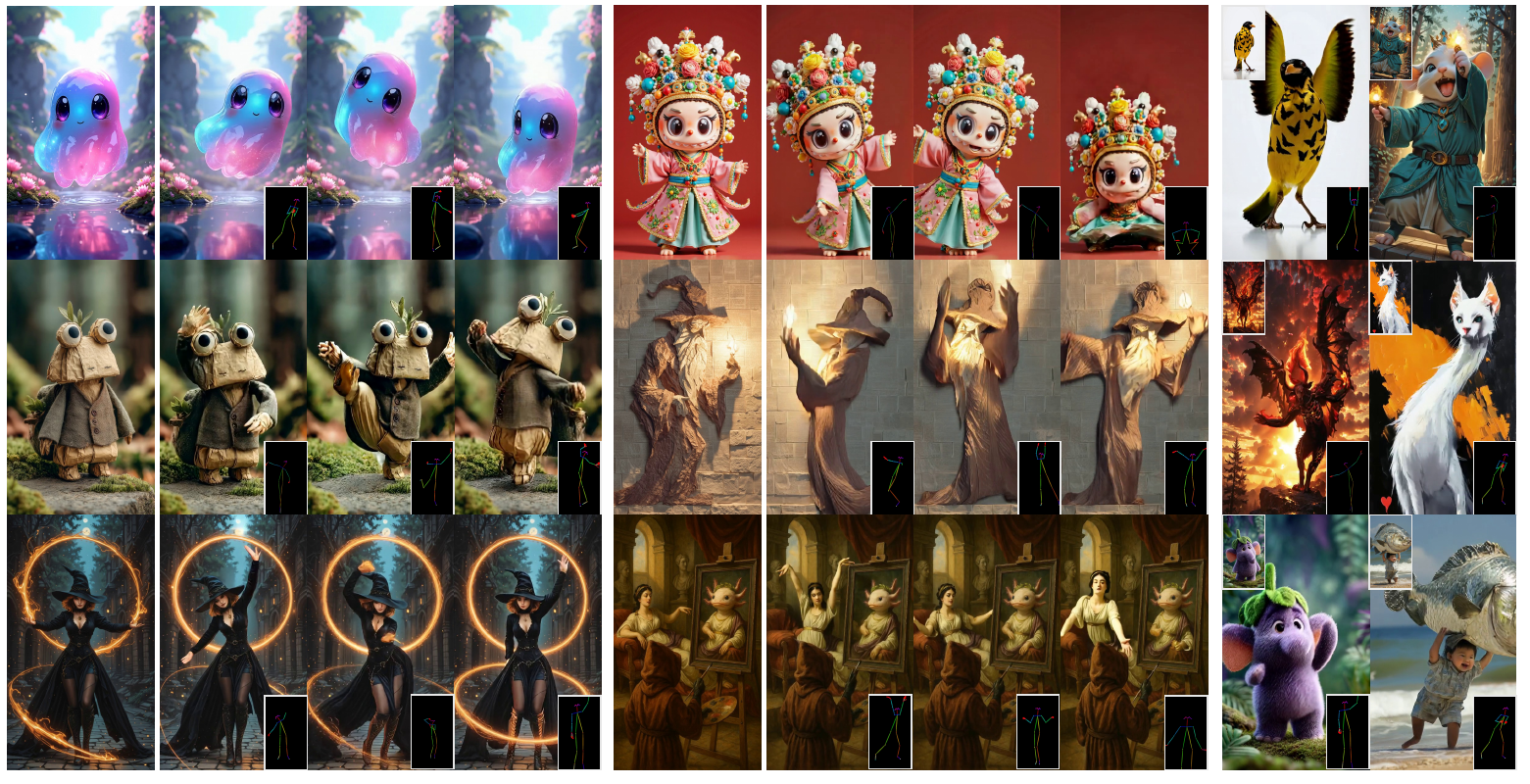}
    \caption{Animations produced by \method which extends beyond human to anthropomorphic characters with various body structures, \textit{e.g.}, without limbs, from games, animations, and posters. Notably, our framework brings the background dynamics to life, such as the formation of a glowing aura, dynamic scene relighting, and the ebb and flow of ocean tides.}
    \vspace{-10pt}
    \label{fig:teaser}
\end{figure*}

The limitations are clearly evidenced for non-human characters in Fig.~\ref{fig:compare}. To replicate the given poses, the diffusion models trained on human dance video datasets tend to introduce unrelated human characteristics which may not make sense to reference figures, resulting in abnormal distortions. In other words, these models treat identity preservation and motion consistency as \textit{conflicting} goals and struggle to balance them, while motion control often prevails. This issue is particularly pronounced for non-human anthropomorphic characters, whose body structures often differ from human anatomy—such as disproportionately large heads or the absence of arms, as shown in Fig.~\ref{fig:teaser}. 
The primary cause is that the motion representations extracted merely from pose conditions are hard to generalize to a broad range of common cartoon characters with unique physical characteristics, leading to their excessive sacrifices in identity preservation in favor of strict pose consistency, which is an unsensible trade-off between these \textit{conflicting} goals. 

To address this issue, the natural approach is to enhance the flexibility of motion representations without discarding current pose condition, which can prevent the model from making unsensible trade-offs between overly precise poses and low fidelity to reference images. To this end, we identify two key limitations of existing methods. \textbf{First}, {the simple 2D pose skeletons}, constructed by connecting sparse keypoints, lack of image-level details and therefore cannot capture the essence of the reference video, such as motion-induced deformations (e.g., body part overlap and occlusion) and overall motion patterns.
\textbf{Second}, the self-driven reconstruction strategy aligns reference and pose skeletons by body shape, simplifying animation but ignoring shape differences during inference. \textbf{Third}, the available animation dataset also typically features static backgrounds and thus provides no supervision for dynamic scene generation.
These inspire us to design the new Pose Indicator from both implicit and explicit perspectives and introduce a new background animation task.

In this paper, we propose \method for animating any character \texttt{X} with dynamic background. Sparked by generative diffusion models~\cite{stablediffusion, wan2025}, we employ a Diffusion Transformer (DiT)~\cite{peebles2023scalable} as the denoising network and provide it with motion feature and figure identity as condition.
To fully capture the gist of motion from the driving video, we introduce the Pose Indicator, which consists of the Implicit Pose Indicator (IPI) and the Explicit Pose Indicator (EPI).
Specifically, IPI extracts implicit motion-related features with the assistance of CLIP image feature, isolating essential motion patterns and relations that cannot be directly represented by the pose skeletons from the driving video. Meanwhile, EPI enhances the representation and understanding of the pose encoder by simulating real-world misalignments between the reference image and driven poses during training, strengthening the ability to generate explicit pose features. With the combined power of implicit and explicit features, \method demonstrates strong character generalization and pose robustness, enabling general \texttt{X} character animation even though it is trained solely on human datasets. 
Furthermore, to address the challenge of static backgrounds, we introduce a multi-task training strategy. Alongside the primary character animation task, we incorporate an auxiliary Text-Image-to-Video (TI2V) task to equip our model with the capability for text-driven background dynamics. To prevent this auxiliary task from degrading the performance of character animation, we employ a partial parameter training strategy, where only a subset of the parameters are updated for the TI2V task. This approach effectively leverages existing character animation and text-to-video datasets, enabling \method to achieve not only high-fidelity character animation but also controllable background dynamics within a single, unified framework.
Moreover, we introduce a new \textbf{A}nimated \textbf{A}nthropomorphic \textbf{Bench}mark (\benchmark), which includes 500 anthropomorphic characters along with corresponding dance videos, to evaluate the performance of \method on other types of characters. Extensive experiments on both public human animation datasets and \benchmark demonstrate that \method outperforms state-of-the-art methods in preserving identity and maintaining motion consistency in animating \texttt{X}. Main contributions summarized as follows:

\begin{itemize}
    \item We present \method, which facilitates image-conditioned pose-guided video generation with high generalizability, particularly for attractive anthropomorphic characters. To the best of our knowledge, this is the first work to animate generic cartoon images without the need for strict pose alignment.
    \item The rethinking about the motion inspire us to propose Pose Indicator, which extracts motion representation suitable for anthropomorphic characters in both implicit and explicit manner, enhancing the robustness of \method. 
    \item We devise a multi-task, partial-parameter training paradigm that seamlessly integrates text-driven background dynamics into the character animation process. This allows for the generation of fully dynamic scenes within a single, unified model.
    \item Since the popular datasets only contain human video with limited character diversity, we present a new \benchmark, specifically for evaluating performance on anthropomorphic characters. Extensive experiments demonstrate that our \method outperforms the competing methods quantitatively and qualitatively on both \benchmark and current human animation benchmark.
\end{itemize}

\textbf{Difference from our conference version.} This manuscript presents significant improvements over our previous conference version~\cite{animatex} including: \textbf{1)} We replace the original 3D Unet with a Diffusion Transformer (DiT) as the backbone, which results in significantly higher video generation quality. Furthermore, we have redesigned the injection methods for the pose condition and the Pose Indicator (IPI \& EPI) to better align with the DiT architecture. \textbf{2)} We introduce a multi-task learning paradigm with a partial-parameter training strategy, enabling our framework to generate dynamic, text-driven backgrounds concurrently with character animation. \textbf{3)} We have significantly expanded the original \benchmark by incorporating a wider variety of character styles to enhance its diversity and complexity. Moreover, the benchmark is now structured into three distinct difficulty levels to facilitate more targeted and systematic evaluation. \textbf{4)} We conduct more comprehensive experiments, further demonstrating the superiority of our method over existing SOTAs on multiple benchmarks.

\section{Related Work}
\label{sec: related_work}

\subsection{Diffusion models for image/video generation}

In recent years, diffusion models~\cite{DDIM,DDPM} have demonstrated strong generative capabilities, pushing image generation technique towards a daily productivity tool~\cite{GLIDE,Dalle2,T2i-adapter,huang2023composer,controlnet,liu2023survey}. Pioneering works such as DALL-E 2~\cite{Dalle2} and Imagen~\cite{saharia2022photorealistic} have showcased the extraordinary potential of diffusion models for high-quality image synthesis. These models employ iterative refinement processes that gradually transform noise into coherent images, enabling the generation of photo-realistic and creatively diverse visuals. 
Notable contributions, including Stable Diffusion~\cite{stablediffusion}, have well balanced scalability and efficiency, making diffusion-based image generation accessible and versatile across various applications. On the video generation front, diffusion models are making amazing progress~\cite{modelscopet2v, tft2v,tune-a-video,chai2023stablevideo,ceylan2023pix2video,guo2023animatediff,zhou2022magicvideo,an2023latent,xing2023simda,qing2023hierarchical,yuan2023instructvideo, tan2024style2talker, gong2024check, wei2024dreamvideo, wei2024dreamvideo2, tan2024mimir, shi2024motionstone,tan2025SynMotion}. These methods joint spatio-temporal modeling to generate realistic motion dynamics and ensure temporal consistency, marking a substantial step forward in generative models for video content. For instance, StableVideo~\cite{chai2023stablevideo} extends the principles of image generation to the temporal domain, enabling the synthesis of coherent and contextually relevant video sequences. 
In this work, we aim to tackle the character-centered image animation task, a dedicated of conditional video generation. Our approach
enables the transformation of static images into dynamic animations by conditioning on desired motion. This innovation bridges the gap between image and video generation, highlights the versatility and adaptability of diffusion models in creating visual narratives.

 \subsection{Pose-guided character motion transfer}
Character image animation aims to transfer motion from the source character to the target identity~\cite{mimicmotion2024, chang2023magicpose, chen2025dancetogether,zhang2025flexiact,tu2025stableanimator++,qu2025evanimate}, which has experienced an impressive journey to improve animation quality and versatility. Early works~\cite{li2019dense, siarohin2019first, siarohin2021motion, zhao2022thin, tan2024edtalk, wang2022latent, tan2024flowvqtalker, tan2024say, tan2023emmn, pan2024expressive,tan2025fixtalk} predominantly utilize Generative Adversarial Networks (GANs) to generate animated human images. However, these GAN-based models are often confronted by the emergence of various artifacts in the generated outputs. With the advent of diffusion models, researchers~\cite{shen2024advancing, zhu2024champ} explored how to go beyond GANs. One effort is Disco~\cite{wang2023disco}, which leverages ControlNet~\cite{zhang2023adding} to facilitate human dance generation, demonstrating the potential of diffusion models in generating dynamic human poses. Following this, MagicAnimate~\cite{xu2023magicanimate} and Animate Anyone~\cite{Animateanyone} introduce transformer-based temporal attention modules~\cite{vaswani2017attention}, enhancing the temporal consistency of animations and resulting in more smooth movement transitions. Building on this, Champ\cite{zhu2024champ} incorporate four distinct conditions to accurately capture the intricate geometry and motion characteristics of humans from source videos. However, maintaining identity consistency, especially in facial regions, remains a persistent challenge that often requires external post-processing. To tackle this, StableAnimator~\cite{tu2025stableanimator} proposes an end-to-end ID-preserving framework. It introduces a distribution-aware ID Adapter to effectively integrate facial identity features while preventing interference from temporal layers, and further employs a Hamilton-Jacobi-Bellman (HJB) equation-based optimization during inference to refine face quality. Sparked by the excellent long sequence processing performance~\cite{Visionmamba,yang2024plainmamba,liu2024vmamba,chen2024video,li2024videomamba} and linear time efficiency of Mamba ~\cite{mamba, gu2021efficiently} conceptually merges the merits of parallelism and non-locality, Unianimate~\cite{wang2024unianimate} resorts to it for efficient temporal modeling. In a similar vein of exploring different architectures, HyperMotion~\cite{xu2025hypermotion} utilizes a Diffusion Transformer (DiT) backbone to address the specific challenge of animating complex human motions. It proposes a Spatial Low-Frequency Enhanced Rotary Position Embedding (SLF-RoPE) to better maintain structural integrity during highly dynamic actions and introduces the Open-HyperMotionX benchmark to support further research.

While these approaches have improved the realism of the animations, a notable limitation remains: most current methods require strict alignment between a reference image and driving video. This restricts their applicability in the scenarios where poses cannot be easily extracted, such as anthropomorphic characters, often resulting in bizarre and unsatisfactory outputs. Furthermore, previous methods are typically confined to generating static backgrounds, which severely limits the realism and immersiveness of the results.
In contrast, our approach adopts a robust and flexible motion representation to mitigate the dependence on pose alignment. This enables the generation of high-quality animations even in cases where previous methods struggle with non-alignable poses. Moreover, we incorporate a TI2V training objective, empowering our model to generate instruction-driven background dynamics. In this manner, our method enhances the versatility and applicability of character image animation across a broad range of contexts (\texttt{X} character) within fully dynamic scenes.

\begin{figure*}[t]
  \centering
  \includegraphics[width=0.98\linewidth]{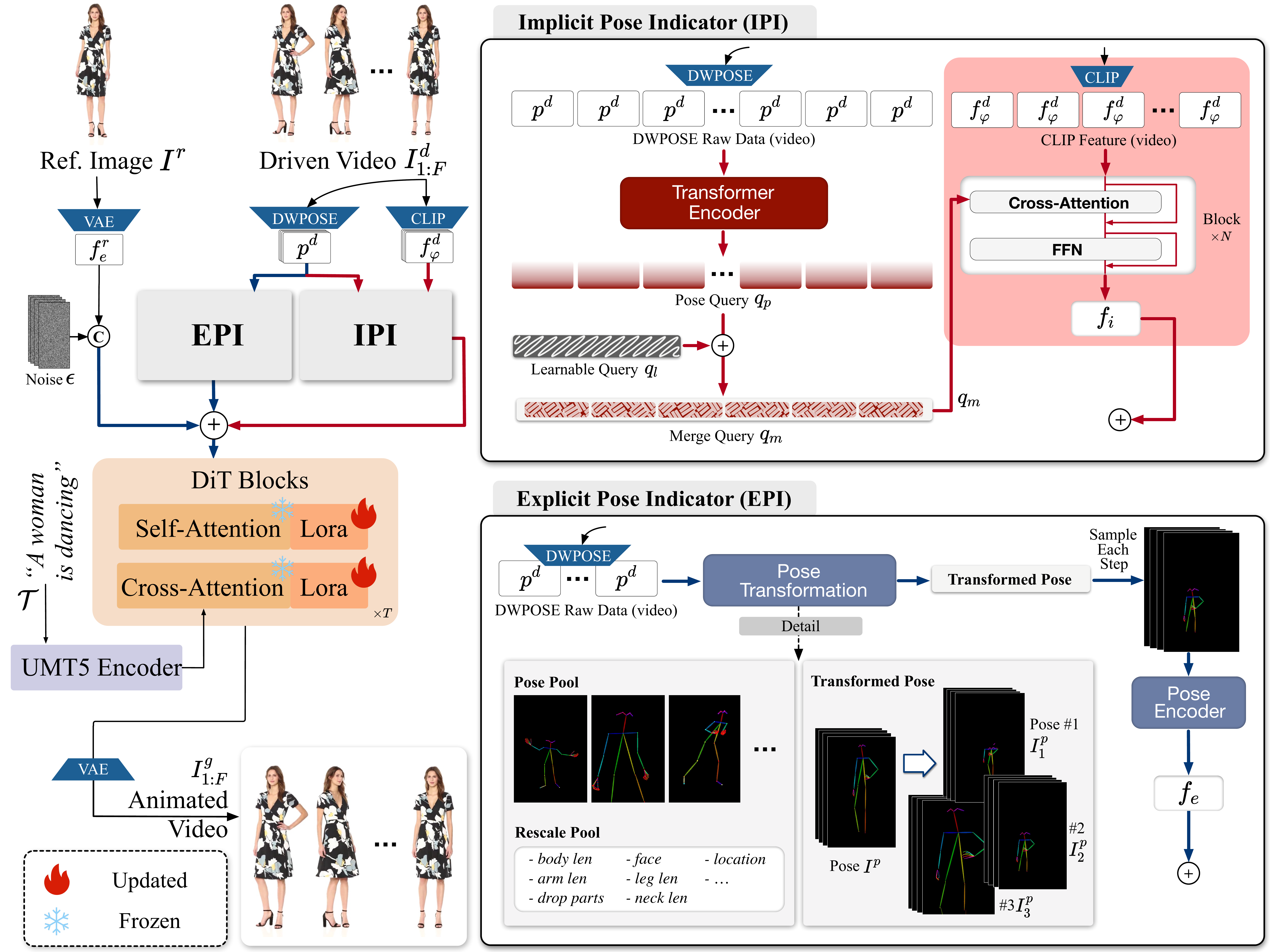}
    \caption{(a) The overview of our \method. Given a reference image $I^r$, we first extract latent feature $f^r_{e}$ via VAE encoder $\mathcal{E}$.
    The proposed Implicit Pose Indicator (\textbf{IPI}) and Explicit Pose Indicator (\textbf{EPI})
    produce motion feature $f_i$ and pose feature $f_e$, respectively. $f^r_e$ is concatenated with the noised input $\epsilon$ along the channel dimension, then further add to $f_e$ and $f_i$. This serves as the input to the diffusion model $\epsilon_\theta$ for progressive denoising. During the denoising process, $\mathcal{T}$ provides background dynamics instruction via a T5 Encoder and cross-attention layer. At last, a VAE decoder $\mathcal{D}$ is adopted to map the generated latent representation $z_0$ to the animation video. (b) The detailed structure of IPT. (c) The pipeline of EPI.}
    \label{fig:method}
    \vspace{-4mm}
\end{figure*}

\section{Method}
\label{sec: method}
In this work, we aim to generate an animated video that
maintains consistency in identity with a reference image $I^r$ and body movement with a driving video $I^d_{1:F}$, while simultaneously generating text-driven dynamics for the background.
Different from previous works, our primary objective is to animate a general characters beyond human, particularly like anthropomorphic ones, which has broader applications.

\subsection{Preliminaries of latent diffusion model}
\label{sec:preliminary}
A diffusion model (DM) operates by learning a probabilistic process that models data generation through noise. To mitigate the heavy computational load of traditional pixel-based diffusion models in high-dimensional RGB spaces, latent diffusion models (LDMs)~\cite{stablediffusion} propose to shift the process into a lower-dimensional latent space using a pre-trained variational autoencoder (VAE)~\cite{kingma2013auto}. It encodes the input data into a compressed latent representation $z_0$. Gaussian noise is then incrementally added to this latent representation over several steps, reducing computational requirements while maintaining the generative capabilities of the model. The process can be formalized as:
\begin{equation}
    q(\mathbf{z}_t \vert \mathbf{z}_{t-1}) = \mathcal{N}(\mathbf{z}_t; \sqrt{1 - \beta_t} \mathbf{z}_{t-1}, \beta_t\mathbf{I}),
\end{equation}
where $\beta_t \in (0, 1)$ represents the noise schedule. As $t \in {1,2,...,T}$ increases, the cumulative noise applied to the original $\mathbf{z}_0$ intensifies, causing $\mathbf{z}_t$ to progressively resemble random Gaussian noise.
Compared to the forward diffusion process, the reverse denoising process $p_\theta$ aims to reconstruct the clean sample $\mathbf{z}_0$ from the noisy input $\mathbf{z}_t$. We represent the denoising step $p(\mathbf{z}_{t-1} \vert \mathbf{z}_t)$ as follows:
\begin{equation}
p_\theta(\mathbf{z}_{t-1} \vert \mathbf{z}_t) = \mathcal{N}(\mathbf{z}_{t-1}; \boldsymbol{\mu}_\theta(\mathbf{z}_t, t), \boldsymbol{\Sigma}_\theta(\mathbf{z}_t, t)),
\end{equation}
in which $\boldsymbol{\mu}_\theta(\mathbf{z}_t, t)$ refers to the estimated target of the reverse diffusion process and the process typically is achieved by a diffusion model $\boldsymbol{\epsilon}_{\theta}$ with the parameters $\theta$.
To model the temporal dimension, the denoising model $\boldsymbol{\epsilon}_{\theta}$ is commonly built on a Diffusion Transformer architecture~\cite{peebles2023scalable} in video generation methods~\cite{wan2025}. Given condition $c$, they use an L2 loss to reduce the difference between the predicted noise and the GT noise during the optimization process:
\begin{equation}
\mathcal{L} = \mathbb{E}_{{\theta}} \Big[\|\boldsymbol{\epsilon} - \boldsymbol{\epsilon}_\theta(\mathbf{z}_t, t, c)\|^2 \Big],
\label{eq:objective}
\end{equation}
once the reversed denoising stage is complete, the predicted clean latent is passed through the VAE decoder to reconstruct the predicted video in pixel space.

\subsection{Pose Indicator}
\label{sec:pose_indicator}

To extract motion representations, previous works typically detect the pose keypoints via DWPose~\cite{DWpose} from the driven video $I^d_{1:F}$ and further visualize them as pose image $I^p$, which are trained using self-driven reconstruction strategy. 
However, it brings several limitations as mentioned in Sec.~\ref{sec: intro}: (1) The sole pose skeletons lack image-level details and are therefore unable to capture the essence of the reference video, such as motion-induced deformations and overall motion patterns. (2) The self-driven reconstruction training strategy naturally aligns the reference and pose images in terms of body shape, which simplifies the animation task by overlooking likely body shape differences between the reference image and the pose image during inference. Both limitations {weaken} the model to develop a deep, holistic motion understanding, leading to \textbf{inadequate} motion representation. To address these issues, we propose Pose Indicator, which consists of Implicit Pose Indicator (IPI) and Explicit Pose Indicator (EPI).

\begin{figure}[t]
  \centering
  \includegraphics[width=1\linewidth]{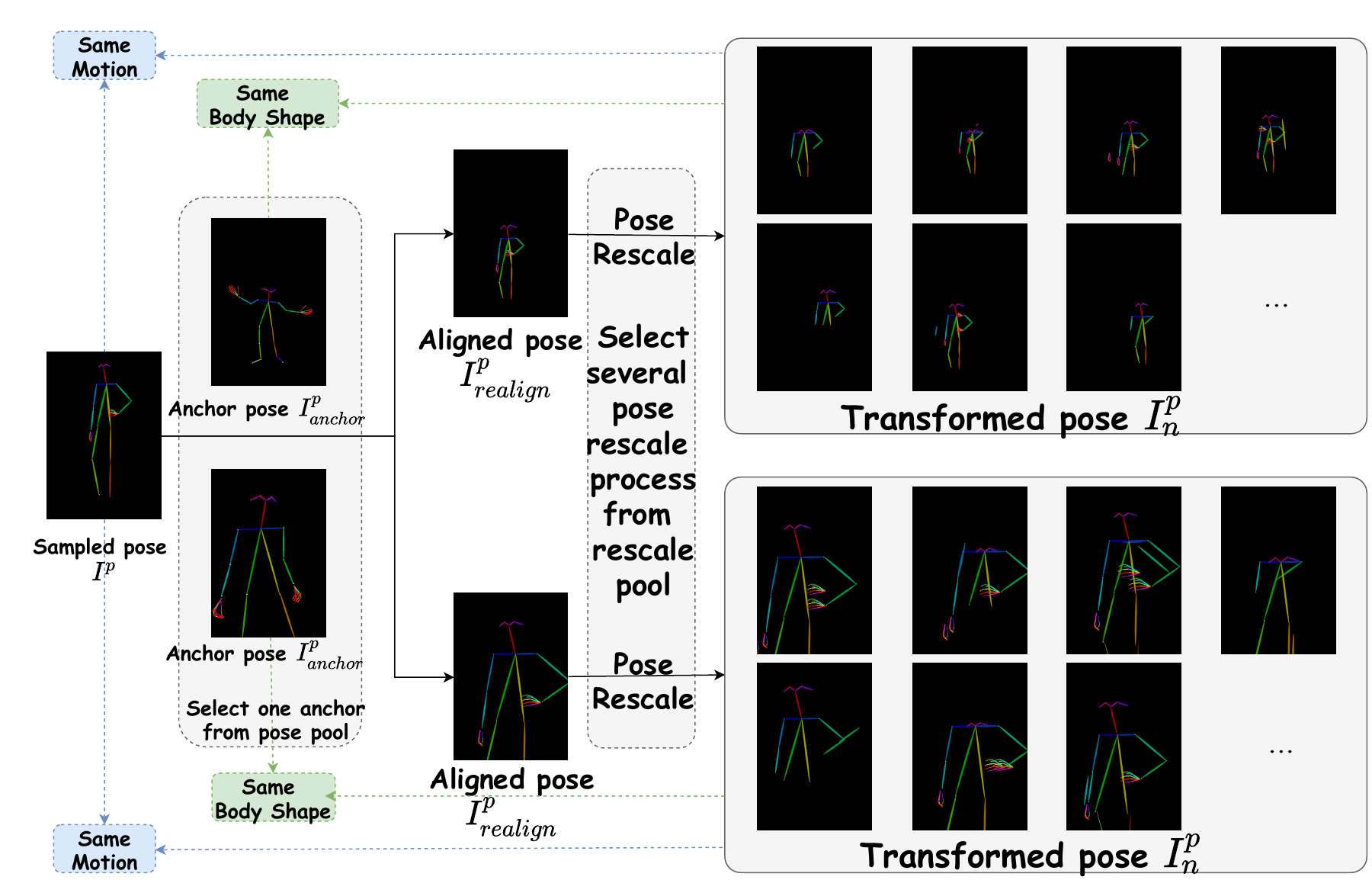}
  \captionsetup{justification=centering} 
    \caption{More example for EPI.}
    \label{fig:more_epi}
    \vspace{-20pt}
\end{figure}

\noindent \textbf{Implicit Pose Indicator (IPI).}
To extract unified motion representations from the driving video in the first limitation, we resort to the CLIP image feature $f^d_\varphi = \Phi(I^d_{1:F})$ extracted by a CLIP Image Encoder. CLIP utilizes contrastive learning to align the embeddings of related images and texts, which may include descriptions of appearance, movement, spatial relationships and etc. Therefore, the CLIP image feature is actually a highly entangled representation, containing motion patterns and relations helpful to animation generation.
As presented in Fig.~\ref{fig:method} (a), we introduce a lightweight extractor $P$ which is composed of $N$ stacked layers of cross-attention and feed-forward networks (FFN). In cross attention layer, we employ $f^d_\varphi$ as the keys (${K}$) and values (${V}$). Consequently, the challenge becomes designing an appropriate query (${Q}$), which should act as a guidance for motion extraction. Considering that the keypoints $p^d$ extracted by DWPose provide a direct description of the motion, we design a transformer-based encoder to obtain the embedding $q_p$, which is regarded as an ideal candidate for ${Q}$. 
Nevertheless, motion modeling using sole sparse keypoints is overly simplistic, resulting in the loss of underlying motion patterns.
To this end, we draw inspiration from query transformer architecture~\cite{awadalla2023openflamingo, jaegle2021perceiver} and initialize a learnable query vector $q_l$ to complement sparse keypoints.
Subsequently, we feed the merged query $q_m = q_p+q_l$ and $f^d_\varphi$ into $P$ and get the implicit pose indicator $f_i$, which contains the essential representation of motion that cannot be represented by the simple 2D pose skeletons.

\noindent \textbf{Explicit Pose Indicator (EPI).}
To deal with the second limitation in the training strategy, 
we propose EPI, designed to train the model to handle misaligned input pairs during inference. 
The \textit{key insight} lies in simulating misalignments between reference image and pose images during training while ensuring the motion remains consistent with the given driving video $I^d_{1:F}$. Therefore, we explore two pose transformation schemes: Pose Realignment and Pose Rescale. As shown in Fig.~\ref{fig:method} (b), in the pose realignment scheme, we first establish a pose pool containing pose images from the training set. In each training step, we first sample the reference image $I^r$ and the driving pose $I^p$ following previous works. Additionally, we randomly select an align anchor pose $I^p_{anchor}$ from the pose pool (two examples are shown in Fig.~\ref{fig:more_epi}). This anchor serves as a reference for aligning the driving pose, producing the aligned pose $I^p_{realign}$. However, since the characters we aim to animate are often anthropomorphic characters, whose shapes can significantly differ from human, such as varying head-to-shoulder ratios, extremely short legs, or even the absence of arms (as shown in Fig.~\ref{fig:teaser} and Fig.~\ref{fig:compare}), relying solely on pose realignment is insufficient to capture these variations for simulation.
Therefore, we further introduce Pose Rescale. Specifically, we define a set of keypoint rescaling operations, including modifying the length of the body, legs, arms, neck, and shoulders, altering face size, even adding or removing specific body parts and etc. These transformations are stored in a rescale pool. After obtaining the realigned poses $I^p_{realign}$, we apply a random selection of transformations from this pool with a certain probability on them, generating the final transformed poses $I^p_n$ (More examples in Fig.~\ref{fig:more_epi}). However, it is important to note that in each training step, only one anchor pose $I^p_{anchor}$ and one rescaling combination are selected, so only one transformed pose $I^p_{n}$ is used for training. As shown in Fig.~\ref{fig:more_epi}, the transformed pose $I^p_{n}$ retains the same motion as the sampled pose $I^p$ but has a body shape similar to the anchor pose $I^p_{anchor}$. This simulates scenarios during inference where there are body shape differences between the reference image and the driving pose, enabling the model to generalize to such cases. We set the probability of $\lambda \in [0,1]$ to apply the pose transformation, and with a probability of $1-\lambda$, the pose image remains unchanged. Subsequently, $I^p_n$ is encoded to the explicit feature $f_e$ via a Pose Encoder.

\subsection{Framework}
\label{sec:Animate-X}
In light of the success of previous works~\cite{unianimatedit}, \method follows the main framework, which consists of several encoders for feature extraction and a DiT~\cite{wan2025, peebles2023scalable} for video generation. As shown in Fig.~\ref{fig:method}, to reduce the parameters of the framework and facilitate appearance alignment, we exclude the Reference Net presented in most of the previous works~\cite{Animateanyone, mimicmotion2024, zhu2024champ}. Instead, a VAE encoder $\mathcal{E}$ is utilized to extract the latent representation $f^r_e$ from reference image $I^r$, which is then directly used as part of the input for the denoising network $\epsilon_\theta$ following~\cite{wan2025, unianimatedit}. For the driven video $I^d_{1:F}$, we detect the pose keypoints $p^d$ and CLIP feature $I^d$ via a DWPose~\cite{DWpose} and CLIP Image Encoder $\Phi$. Subsequently, IPI and EPI introduced in Sec.~\ref{sec:pose_indicator} extract the implicit latent $f_i$ and explicit latent $f_e$, respectively. The $f^r_e$ is first concatenated with the noised latent $\epsilon$ to obtain the fused features along the channel dimension, which is further added to $f_e$ and $f_i$, resulting in combined features $f_{merge}$. Then, the combined features are fed into the video diffusion model $\epsilon_\theta$ for jointly appearance alignment and motion modeling. The diffusion model $\epsilon_\theta$ comprises multiple stacked DiT blocks, each of which consists of Self-Attention and Cross-Attention. The Self-Attention receives inputs from $f_{merge}$ and fuses the identity condition from $I^r$ with the motion condition from $I^d$, producing an intermediate representation $x$. Subsequently, the prompt $\mathcal{T}$ is fed into the Cross-Attention module along with $x$, enabling instruction control. For efficient adaptation, we fine-tune only a small subset of parameters using Low-Rank Adaptation (LoRA), which reduces memory overhead and enhances adaptability without compromising performance. 

\begin{figure}[t]
  \centering
  \includegraphics[width=\linewidth]{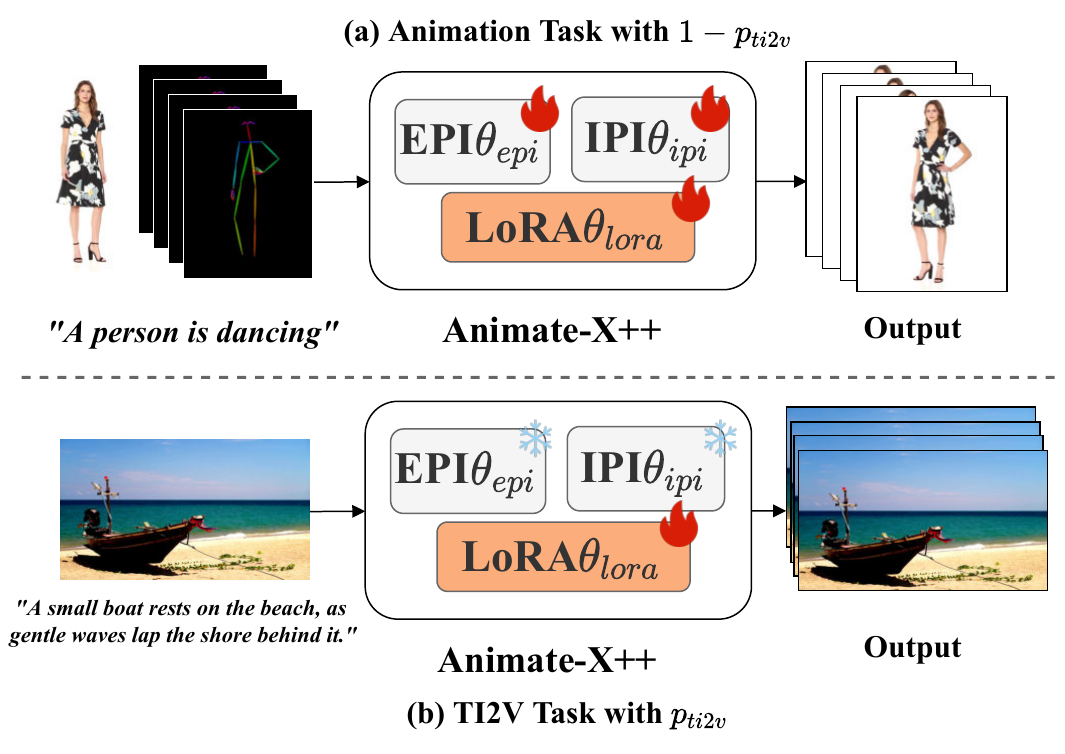}
  \captionsetup{justification=centering} 
    \caption{Multi-task and partial parameter training strategy.}
    \label{fig:multitask}
    \vspace{-15pt}
\end{figure}

\subsection{Multi-Task Training for Dynamic Backgrounds}
While the Pose Indicator provides a robust motion representation for general character animation, the generated videos are still confined to static backgrounds. To address this, we aim to enable dynamic backgrounds guided by a text prompt, which controls scene dynamics beyond the foreground character. However, this is challenging for two main reasons: (1) existing character animation datasets typically feature static backgrounds and lack corresponding detailed text descriptions, and (2) naively integrating a new task for background dynamics risks degrading the model's primary performance on character animation.

To tackle these issues, we devise a multi-task training framework that jointly learns character animation and a Text-Image-to-Video (TI2V) task. Specifically, we augment our training animation data $D_{\text{anim}}$ with a text-video dataset $D_{\text{ti2v}}$. As illustrated in Fig.~\ref{fig:multitask}, at each training step, we sample from the animation dataset with a probability of $1-p_{ti2v}$ to train the character animation task, and from the text-video dataset with a probability of $p_{ti2v}$ to train the TI2V task. Notably, despite the data from both tasks featuring varying aspect ratios (e.g., landscape and portrait), thanks to the Patchify mechanism of the DiT backbone and the use of 3D-RoPE~\cite{su2024roformer} for relative positional encoding, our method supports training on variable resolutions simultaneously. However, we find that directly mixing these tasks led to a severe drop in animation quality. This is primarily because the TI2V task lacks pose guidance, causing the model to weaken its meticulously trained pose-alignment capabilities. To resolve this, we introduce a partial parameter training strategy. When training on the character animation task, we update all learnable parameters of the model. In contrast, when training on the TI2V task, we set the output of Pose Indicator as 0 and freeze all pose-related modules (\textit{i.e.,} IPI $\theta_{\text{ipi}}$, and EPI $\theta_{\text{epi}}$), and only update a set of lightweight LoRA parameters $\theta_{\text{lora}}$ injected into the model:
\begin{equation}
\label{eq:total_loss}
L_{\text{total}} = \mathbb{E}_{(I^r, I^p_{1:F}) \sim D_{\text{anim}}}[L_{\text{anim}}] + \mathbb{E}_{(I^s, c_t) \sim D_{\text{ti2v}}}[L_{\text{ti2v}}],
\end{equation}
\begin{equation}
L_{\text{anim}} = L_{\text{diff}}(I^r, I^d_{1:F};\, \theta_{\text{ipi}},\theta_{\text{epi}},\theta_{\text{lora}})
\end{equation}
\begin{equation}
L_{\text{ti2v}} = L_{\text{diff}}(I^r, \mathcal{T};\, \text{stop\_grad}(\theta_{\text{ipi}},\theta_{\text{epi}}), \theta_{\text{lora}})
\end{equation}

This strategic approach allows us to successfully co-train both tasks, enabling the model to learn rich, text-driven background dynamics without compromising the performance of its specialized character animation components.
\begin{figure}[t]
  \centering
  \includegraphics[width=0.95\linewidth]{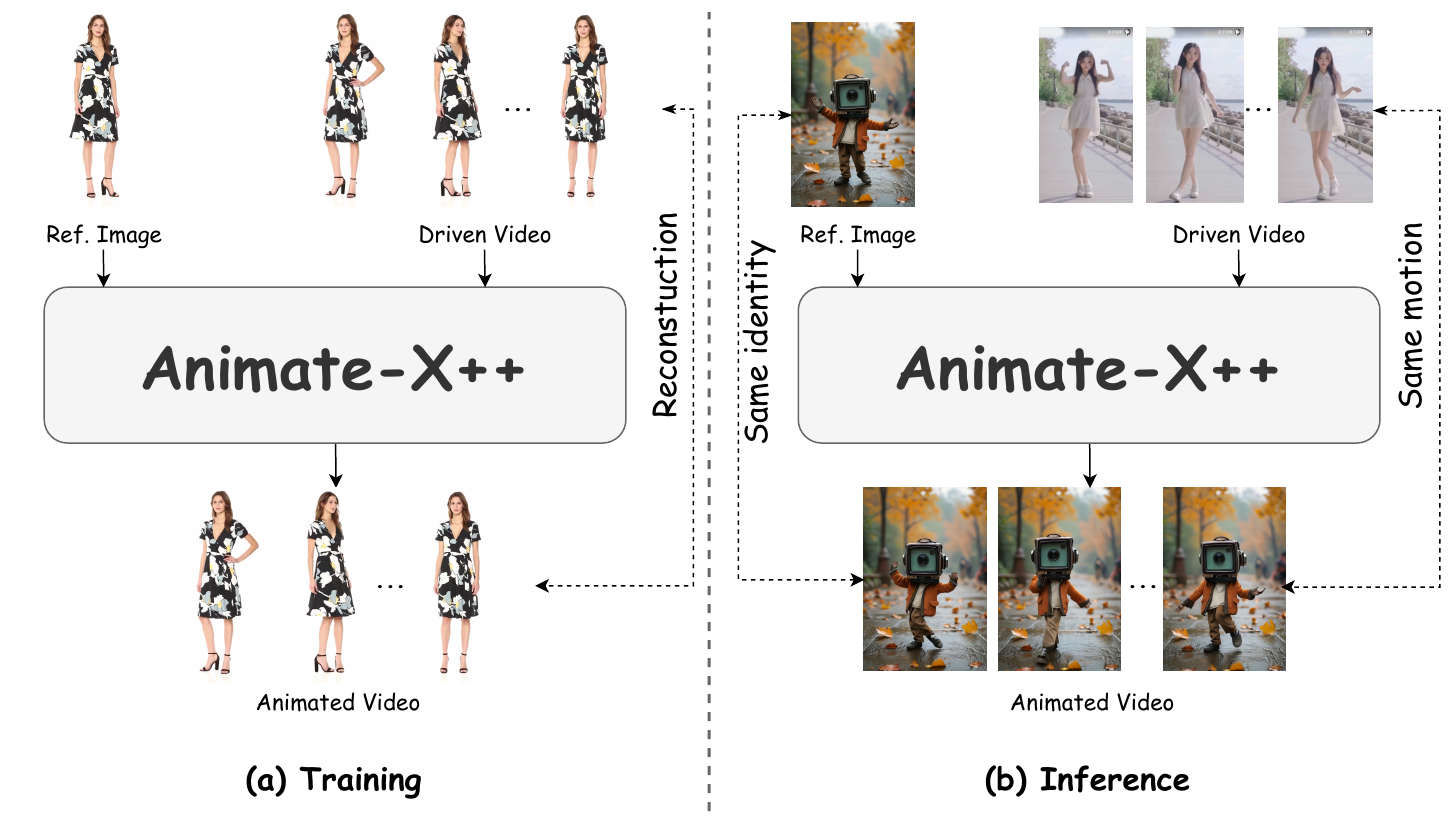}
  \captionsetup{justification=centering} 
    \caption{The difference of training and inference pipeline.}
    \label{fig:train_inference}
\end{figure}

\noindent \textbf{Training and Inference.} 
To improve the model's robustness against pose and reference image misalignments, we adopt two key training schemes. First, we set a high transformation probability $\lambda$ (over 98\%) in the EPI, enabling the model to handle a wide range of misalignment scenarios. Second, we apply random dropout to the input conditions at a predefined rate~\cite{wang2024unianimate}. Third, we assign a relatively low probability of $p_{ti2v}$ to the TI2V task, thereby prioritizing the learning of character animation as the main objective. After that, while the reference image and driven video are from the same human dancing video during training, in the inference phase (Fig.~\ref{fig:train_inference} (b)), \method can handle an arbitrary reference image and driven video, which may differ in appearance.

\begin{figure*}[t]
  \centering
  \includegraphics[width=1\linewidth]{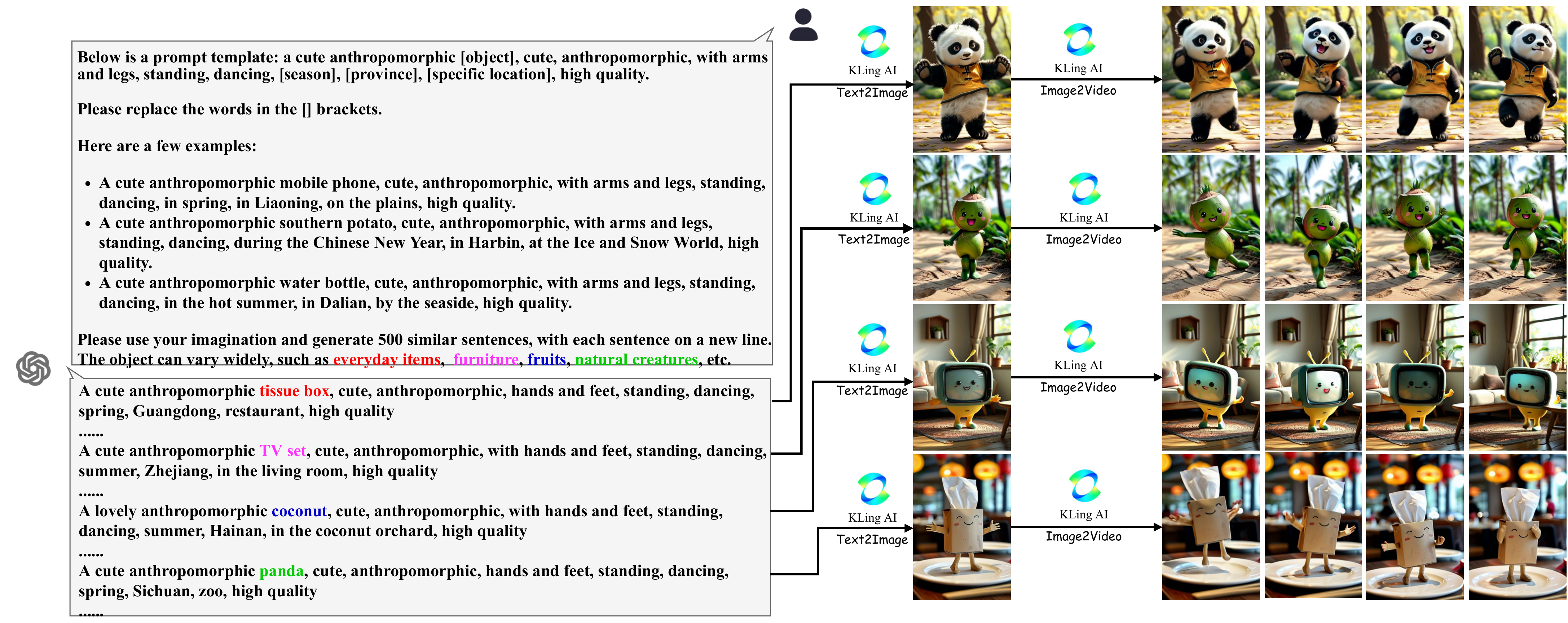}
  \captionsetup{justification=centering} 
    \caption{Detailed pipeline for building \benchmark.}
    \label{fig:data_prepare}
    \vspace{-4mm}
\end{figure*}

\begin{figure}[t]
  \centering
  \includegraphics[width=\linewidth]{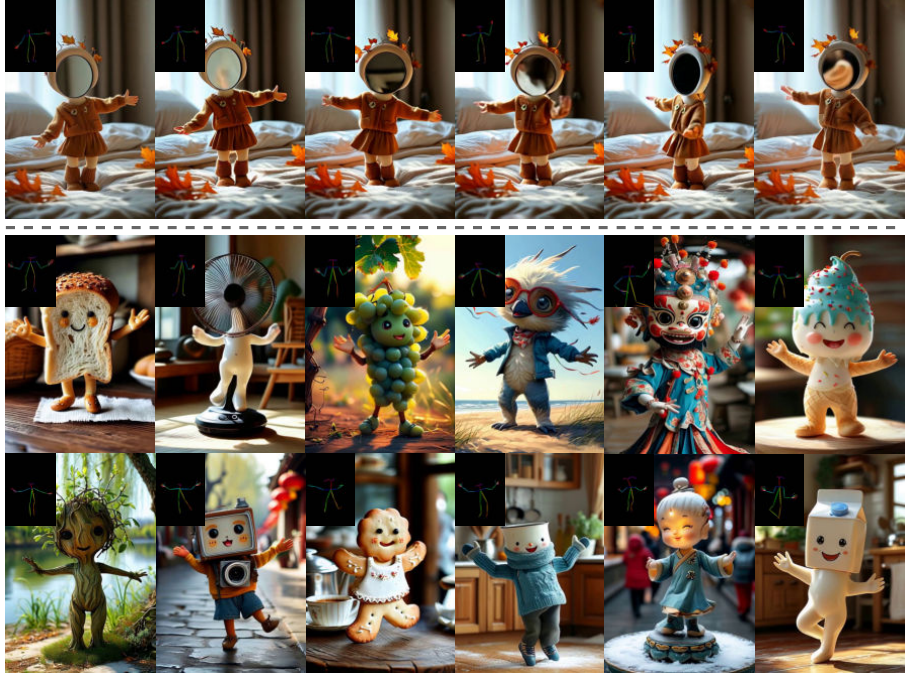}
\captionsetup{justification=centering} 
    \caption{Examples from our \benchmark.}
    \label{fig:benchmark}
\vspace{-4mm}
\end{figure}

\begin{figure}[t]
  \includegraphics[width=\linewidth]{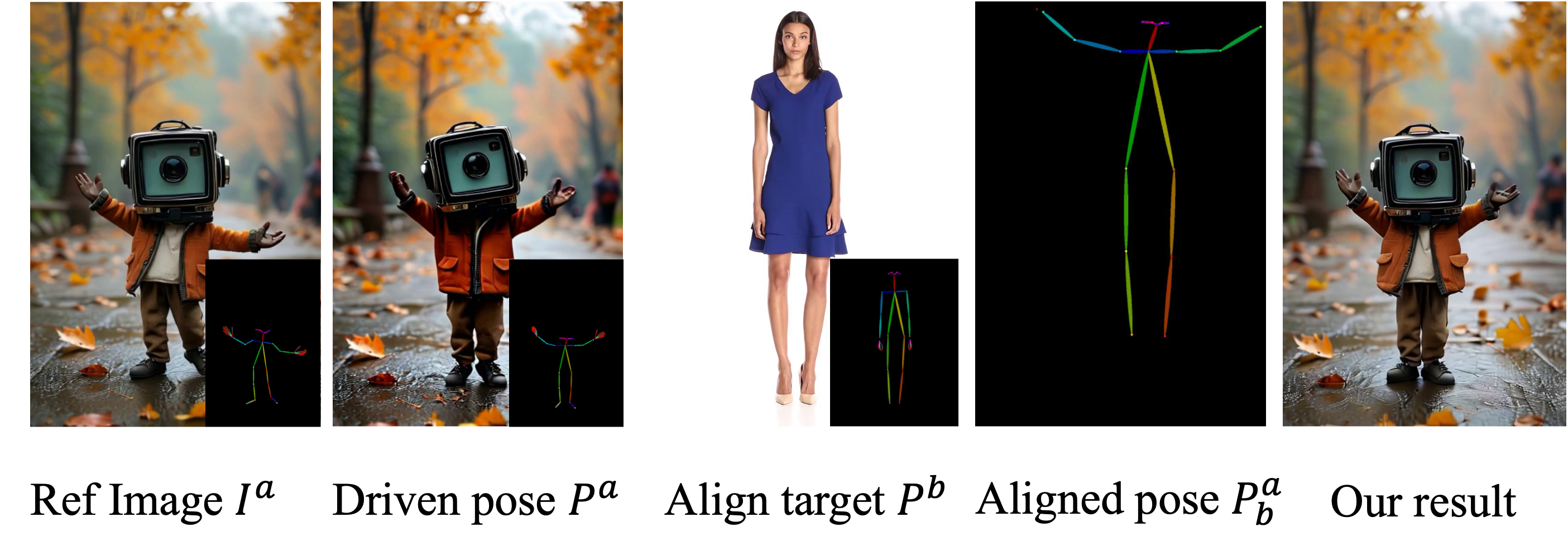}
\captionsetup{justification=centering} 
    \caption{The illustration of comparison settings.}
    \label{fig:setting_demo}
\vspace{-5mm}
\end{figure}

\begin{figure}[t]
  \centering
  \includegraphics[width=\linewidth]{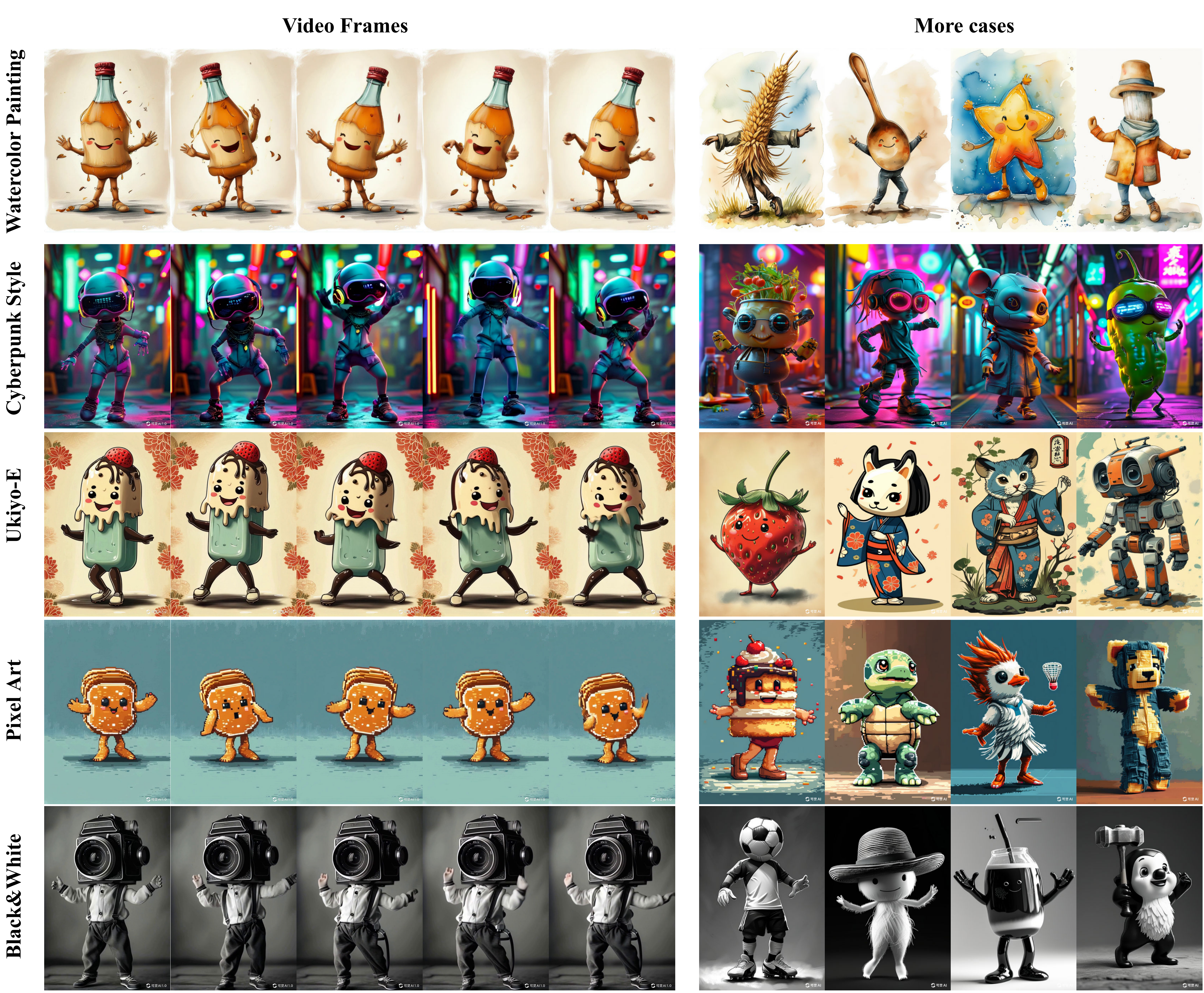}
  \captionsetup{justification=centering} 
    \caption{More styles in $A^2Bench$.}
    \label{fig:more_style}
\vspace{-4mm}
\end{figure}

\subsection{\benchmark}
\label{subsec:benchmark}

The main task of our \method is to animate an anthropomorphic character with vivid and smooth motions. However, current publicly available datasets~\cite{TikTokdata, UBCfashion} primarily focus on human animation and fall short in capturing a broad range of anthropomorphic characters and corresponding dancing videos. This gap makes these datasets and benchmarks unsuitable for quantitatively evaluating different methods in anthropomorphic character animation. 

To bridge this gap, we propose the \textbf{A}nimated \textbf{A}nthropomorphic character \textbf{Bench}mark (\benchmark) to comprehensively evaluate the performance of different methods. As shown in Fig.~\ref{fig:data_prepare}, initially provide GPT-4o with a template that clearly specifies the demand to generate `anthropomorphized' images. The images were required to be cute, with arms and legs, standing, dancing, and of high quality. To allow for a variety of image outputs, we left the fields for `object', `season', `province', and `specific location' empty. For the key factor influencing diversity and relevance, i.e., `object', we provide a selectable range, such as everyday items, furniture, fruits, and natural creatures. To help GPT-4o better understand our intent, we additionally provide two examples, where the prompts had already been proven to generate satisfactory images by text-to-image module of KLing AI. Thanks to the text understanding and generation capabilities of GPT-4o, we collect 500 prompts for image generation. We then fed these 500 prompts into the text-to-image module of Keling AI, obtaining corresponding anthropomorphic characters images. Based on these images, we further generate videos of them dancing using the image-to-video module of Keling AI. In this way, we collect 500 pairs of images and videos of anthropomorphic characters, forming our \benchmark, as shown in Fig.~\ref{fig:benchmark}. Moreover, we add style trigger words such as ``\textit{Watercolor Painting}'', ``\textit{Cyberpunk Style}'', ``\textit{Van Gogh}'', ``\textit{Ukiyo-E}'', ``\textit{Pixel Art}'' and so on. The results are presented in Fig.~\ref{fig:more_style}, which further enhances the diversity and complexity of \benchmark.

Since most current animation methods~\cite{wang2024unianimate, Animateanyone, mimicmotion2024} take a pose image sequence as motion source, we also provide our \benchmark with additional pose images. To achieve this, we employ DWPose~\cite{DWpose} to extract pose sequences from the videos. However, since DWPose is trained on human data, it does not accurately extract every pose in the dancing video of the anthropomorphic character, so after extraction, we manually screen 100 videos with accurate poses, and view them as test videos for calculating quantitative metrics. Fig.~\ref{fig:benchmark} displays several examples, which include anthropomorphic characters of plants, animals, food, furniture, etc. For images and videos where pose extraction is not feasible, we take them as key sources of reference images in our qualitative demonstrations. This will inspire the community to animate a wider range of interesting cases. We also anticipate that these data could serve as an important resource for future pose extraction algorithms tailored to anthropomorphic datasets, making them accessible for broader use.

\begin{table*}[!t]
\caption{
Quantitative comparisons with SOTAs on \benchmark with the rescaled pose setting.
``PSNR*'' means using the modified metric~\cite{disco} to avoid numerical overflow.}
\renewcommand{\arraystretch}{1}
\setlength\tabcolsep{2.5pt}
\centering
\resizebox{0.85\linewidth}{!}{
\begin{tabular}{l|ccccc|cc}
\shline
Method         & PSNR* $\uparrow$ & SSIM $\uparrow$ & L1 $\uparrow$ & LPIPS $\downarrow$ & FID $\downarrow$  &FID-VID $\downarrow$  & FVD $\downarrow$ \\ \shline
Moore-AnimateAnyone~\cite{Moore}   & 9.86  & 0.299 & 1.58E-04 & 0.626 & 50.97  & 75.11  & 1367.84         \\
MimicMotion~\cite{mimicmotion2024} $_{\color{gray}{\text{(ICML25)}}}$    & 10.18 & 0.318 & 1.51E-04 & 0.622 & 122.92 & 129.40 & 2250.13        \\
ControlNeXt~\cite{peng2024controlnext} $_{\color{gray}{\text{(ArXiv24)}}}$    & 10.88 & 0.379 & 1.38E-04 & 0.572 & 68.15  & 81.05  & 1652.09     \\
MusePose~\cite{musepose} $_{\color{gray}{\text{(ArXiv24)}}}$   & 11.05 & 0.397 & 1.27E-04 & 0.549 & 100.91 & 114.15 & 1760.46        \\

Unianimate~\cite{wang2024unianimate} $_{\color{gray}{\text{(SCIS25)}}}$          & {11.82} & {0.398}          & {1.24E-04} & {0.532} & {48.47} & {61.03} & {1156.36}      \\

StableAnimator~\cite{tu2025stableanimator} $_{\color{gray}{\text{(CVPR25)}}}$          & {12.55} & {0.406} & {1.19E-04} & {0.528} & {50.85} & {50.91} & {989.12}      \\

Unianimate-DiT~\cite{unianimatedit} $_{\color{gray}{\text{(ArXiv24)}}}$          & {12.12} & {0.404} & {1.15E-04} & {0.514} & {44.19} & {52.55} & {945.60}     \\

\rowcolor{Gray}

\textbf{Ours-3D Unet~\cite{animatex} $_{\color{gray}{\text{(ICLR25)}}}$}  & \textbf{13.60} & \textbf{0.452} &\textbf{ 1.02E-04} & \textbf{0.430} & \textbf{26.11} & \textbf{32.23} & \textbf{703.87}   \\
\rowcolor{Gray}

\textbf{Ours-DiT(\method)}  & \textbf{13.94} & \textbf{0.462} &\textbf{ 9.34E-05} & \textbf{0.421} & \textbf{25.14} & \textbf{31.09} & \textbf{681.42}   \\

\shline
\end{tabular} 
}

\label{tab:quantitative_cartoon_cross_setting}
\vspace{-1mm}
\end{table*}

\section{Experiments}
\label{sec: exp}

\subsection{Experimental Settings}

\noindent \textbf{Dataset.} For the animation task, we collect approximately 9,000 human videos from the internet and supplement this with the TikTok dataset~\cite{TikTokdata} and Fashion dataset~\cite{UBCfashion} for training. In addition, we collect 10,000 text-video pairs for training the TI2V task.
Following previous works~\cite{Animateanyone,UBCfashion,TikTokdata}, we use 10 and 100 videos for both qualitative and quantitative comparisons from TikTok and Fashion dataset, respectively. 
We additionally experimented on 100 image-video pairs selected from the newly proposed \benchmark introduced in Sec~\ref{subsec:benchmark}. Please note that, to ensure a fair comparison, the data in the \benchmark are \textbf{not} included in the training set to train our model. The data are only used to evaluate the quantitative results and provide interesting reference image cases. 

\noindent \textbf{Implement Details.} 
In the experiments, we use the visual encoder of the multi-modal CLIP-Huge model~\cite{CLIP} in Stable Diffusion v2.1~\cite{stablediffusion} to encode the CLIP embedding of the driving videos. The pose encoder, composed of multiple stacked 3D convolutional layers to extract temporal and spatial features of driving poses effectively, follows a similar structure in~\cite{unianimatedit}. For model initialization, we employ a pre-trained WanX2.1~\cite{wan2025}, as done in previous approaches~\cite{unianimatedit}. The experiments are carried out using 8 NVIDIA H20 GPUs. During training, videos in animation datasets are resized to a spatial resolution of $832\times480$ pixels, while for the text-video dataset, they are resized to $480\times832$. We feed the model with uniformly sampled video segments of 81 frames to ensure temporal consistency. We use the AdamW optimizer~\cite{loshchilov2017AdamW} with learning rates of 1e-7 for the implicit pose indicator and 1e-5 for other modules. For noise sampling, DDPM~\cite{DDPM} with 1000 steps is applied during training. In the inference phase, we use the DDIM sampler~\cite{DDIM} with 50 steps for faster sampling. Furthermore, for classifier-free guidance (CFG)~\cite{ho2022classifier}, we set the scale to 5 to achieve a strong balance between animation fidelity and identity preservation, yielding high-quality results.

\begin{table*}[!t]
\caption{
Quantitative comparisons with existing methods on \benchmark in the self-driven setting.}
\renewcommand{\arraystretch}{1}
\setlength\tabcolsep{2.5pt}
\centering
\resizebox{0.85\linewidth}{!}{
\begin{tabular}{l|ccccc|cc}
\shline
Method         & PSNR* $\uparrow$ & SSIM $\uparrow$ & L1 $\uparrow$ & LPIPS $\downarrow$ & FID $\downarrow$  &FID-VID $\downarrow$  & FVD $\downarrow$ \\ \shline

FOMM~\cite{FOMM} $_{\color{gray}{\text{(NeurIPS19)}}}$   & 10.49 & 0.363 & 1.47E-04  & 0.613 & 183.18 & 147.82 & 2535.12         \\
MRAA~\cite{MRAA} $_{\color{gray}{\text{(CVPR21)}}}$    & 12.62 & 0.420 & 1.09E-04  & 0.556 & 161.57 & 196.87 & 3094.68      \\
LIA~\cite{wang2022latent} $_{\color{gray}{\text{(ICLR22)}}}$    & {13.78} & {0.445} & {9.70E-05} & 0.497 & 105.13 & 78.51  & 1813.28    \\
\midrule
DreamPose~\cite{karras2023dreampose} $_{\color{gray}{\text{(ICCV23)}}}$ & 7.76  & 0.305 & 2.28E-04  & 0.534 & 277.64 & 315.58 & 4324.42\\

MagicAnimate~\cite{magicanimate} $_{\color{gray}{\text{(CVPR24)}}}$   &11.90 & 0.396 & 1.17E-04  & 0.523 & 117.09 & 117.54 & 2021.93         \\
Moore-AnimateAnyone~\cite{Moore}  $_{\color{gray}{\text{(CVPR24)}}}$    & 11.56 & 0.360 & 1.27E-04  & 0.532 & {37.82}  & 59.80  & {1117.29}    \\
MimicMotion~\cite{mimicmotion2024} $_{\color{gray}{\text{(ICML25)}}}$   & 12.66 & 0.407 & 1.07E-04  & 0.497 & 96.46  & 61.77  & 1368.83    \\

ControlNeXt~\cite{peng2024controlnext} $_{\color{gray}{\text{(ArXiv24)}}}$    & 12.82 & 0.421 & 1.02E-04  & 0.472 & 46.66  &{ 59.41}  & 1152.96    \\
MusePose~\cite{musepose} $_{\color{gray}{\text{(ArXiv24)}}}$   & 12.92 & 0.438 & 9.90E-05 & {0.470} & 80.22  & 87.97  & 1401.96      \\

StableAnimator~\cite{tu2025stableanimator} $_{\color{gray}{\text{(CVPR25)}}}$          & {13.15} & {0.435} & {9.55E-05} & {0.478} & {60.12} & {78.20} & {1215.50}      \\

Unianimate-DiT~\cite{unianimatedit} $_{\color{gray}{\text{(ArXiv24)}}}$          & {13.52} & {0.444} & {9.21E-05} & {0.461} & {55.67} & {69.88} & {1120.75}     \\

\rowcolor{Gray}

\textbf{Ours-3D Unet~\cite{animatex} $_{\color{gray}{\text{(ICLR25)}}}$}  & \textbf{14.10}       & \textbf{0.463}    & \textbf{8.92E-05}             & \textbf{0.425}            & \textbf{31.58} & \textbf{33.15}                                          &\textbf{849.19}      \\

\rowcolor{Gray}
\textbf{Ours-DiT(\method)}  & \textbf{14.48} & \textbf{0.479} & \textbf{8.51E-05} & \textbf{0.410} & \textbf{28.93} & \textbf{30.05} & \textbf{798.42}         \\
\shline
\end{tabular} 
}

\vspace{-4mm}
\label{tab:quantitative_cartoon_self_setting}
\end{table*}

\noindent \textbf{Evaluation Metrics.}
We employ several evaluation metrics to quantitatively assess our results, including PSNR~\cite{hore2010image}, SSIM~\cite{wang2004image}, L1, LPIPS~\cite{zhang2018unreasonable}, FID~\cite{heusel2017gans}, FID-VID~\cite{balaji2019conditional} and FVD~\cite{unterthiner2018towards}. The detailed metrics are introduced as follows: (1) PSNR is a measure used to evaluate the quality of reconstructed images compared to the original ones. (2) SSIM assesses the similarity between two images based on their luminance, contrast, and structural information. (3) The L1 metric refers to the mean absolute difference between the corresponding pixel values of two images. (4) LPIPS is a perceptual distance metric based on deep learning. It evaluates the similarity between images by analyzing the feature representations of image patches and tends to align well with human visual perception. (5) FID is used to assess the quality of generated images by comparing the distribution of generated images to that of real images in feature space. (6) FID-VID extends the FID metric to video data. It measures the quality of generated videos by comparing the distribution of generated video features to real video features. (7) FVD is another metric for evaluating video generation, which measures the distance between the feature distributions of real and generated videos, taking both spatial and temporal dimensions into account.

\begin{figure*}[t]
  \centering
  \includegraphics[width=\linewidth]{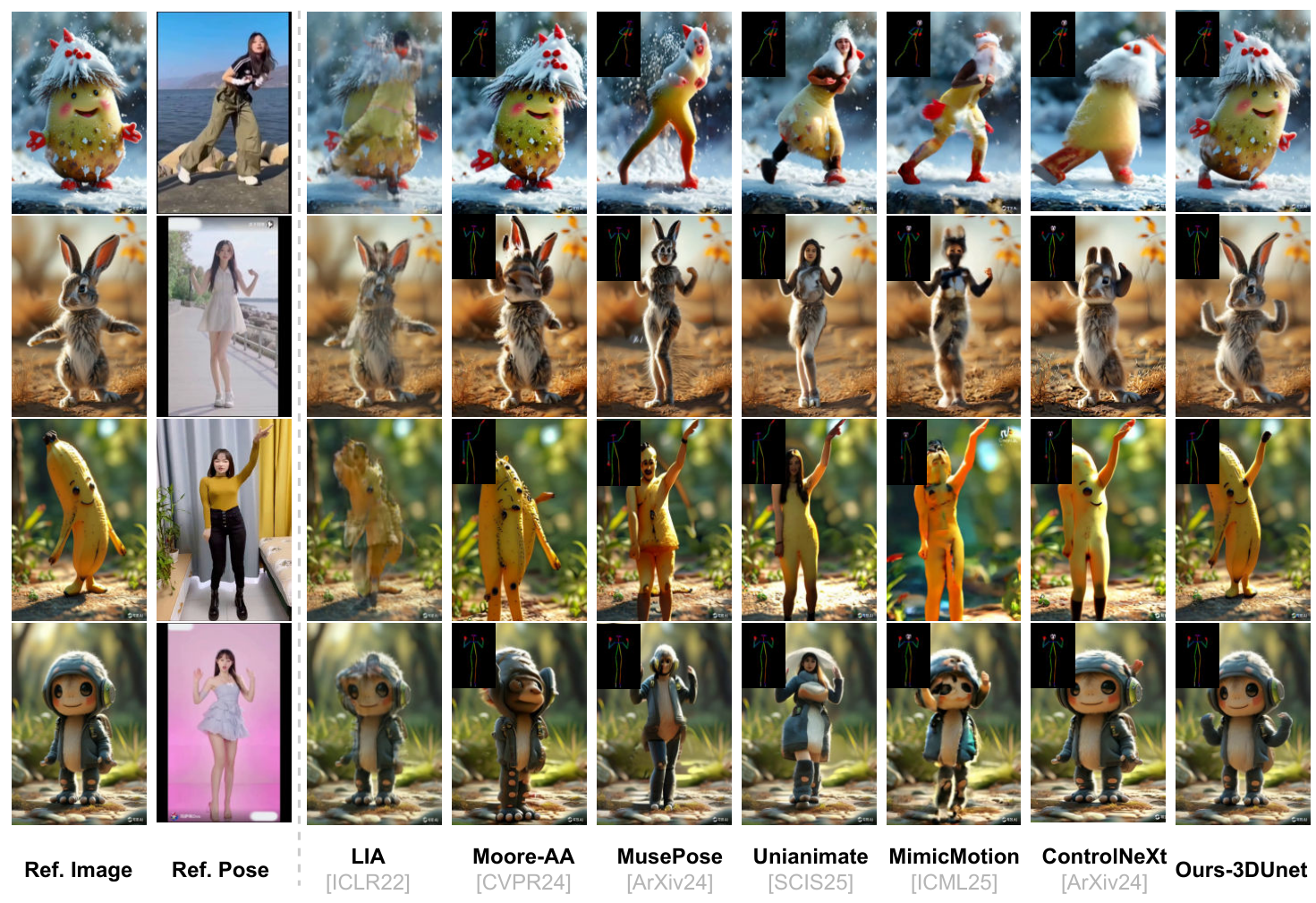}
  \captionsetup{justification=centering} 
    \caption{Qualitative comparisons with state-of-the-art methods.}
    \label{fig:compare}
\vspace{-4mm}
\end{figure*}

\begin{figure*}[t]
  \centering
  \includegraphics[width=0.8\linewidth]{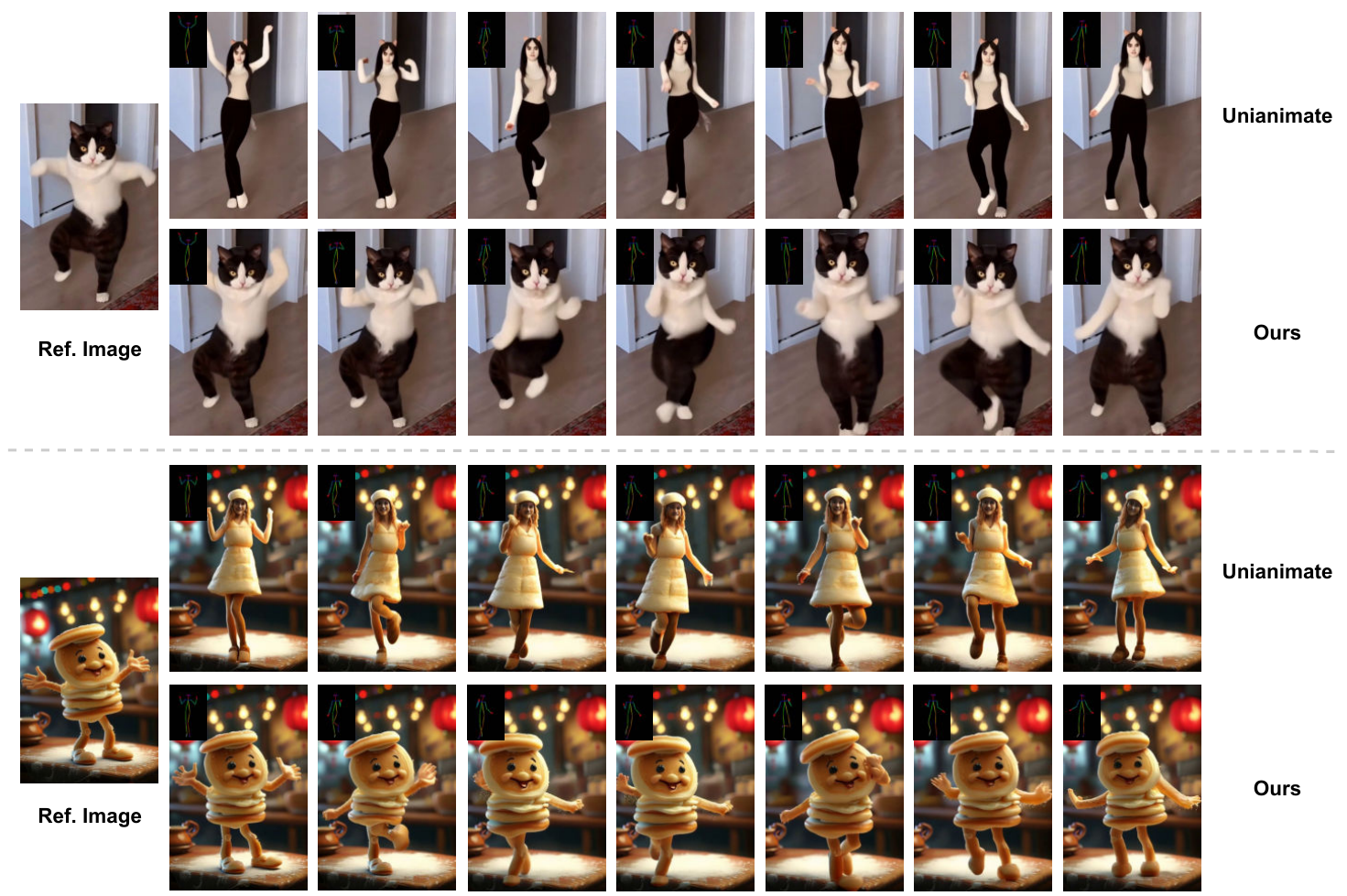}
\captionsetup{justification=centering} 
    \caption{Qualitative comparisons with Unianimate in terms of long video generation.}
    \label{fig:compare_2}
\vspace{-4mm}
\end{figure*}

\subsection{Experimental Results}
\noindent \textbf{Quantitative Results.} Since our \method primarily focuses on animating the anthropomorphic characters, very few of which, if not none, can be extracted the pose skeleton accurately by DWPose~\cite{DWpose}. It naturally leads to a misalignment of the input reference image with the driving pose images. To compute quantitative results in this case, we set up a new comparison setting.
For each case in \benchmark (\textit{i.e.}, a reference image $I^a$ and a pose $P^a$, as shown in Fig.~\ref{fig:setting_demo}), we randomly select one human's pose image $P^b$ and align the anthropomorphic character's pose $P^a$ to it, such that the aligned pose $p^a_b$ retains the movements of $P^a$ but has the same body shape (fat/thin, tall/short, \textit{etc.}) as $p^b$. Ultimately, we take the anthropomorphic character $I_a$ and the aligned driving pose image $p^a_b$ as inputs to the model, generating results that allow it to calculate quantitative metrics with the original anthropomorphic character dancing video in \benchmark. In this setting, we compare our method with Animate Anyone~\cite{Animateanyone}, Unianimate~\cite{wang2024unianimate}, MimicMotion~\cite{mimicmotion2024}, ControlNeXt~\cite{peng2024controlnext}, MusePose~\cite{musepose}, StableAnimator~\cite{tu2025stableanimator} and Unianimate-DiT~\cite{unianimatedit}, which also use pose images (\textit{e.g.,} $P^b$ in Fig.~\ref{fig:setting_demo}) as input. The results of Animate Anyone~\cite{Animateanyone} are obtained by leveraging the publicly available reproduced code~\cite{Moore}. Tab.~\ref{tab:quantitative_cartoon_cross_setting} presents the quantitative results, where \method markedly surpasses all comparative methods in terms of all metrics.
It is worth noting that, we do not use \benchmark as training data to avoid overfitting and ensure fair comparisons, in line with other comparative methods.

\begin{figure*}[t]
  \centering
  \includegraphics[width=\linewidth]{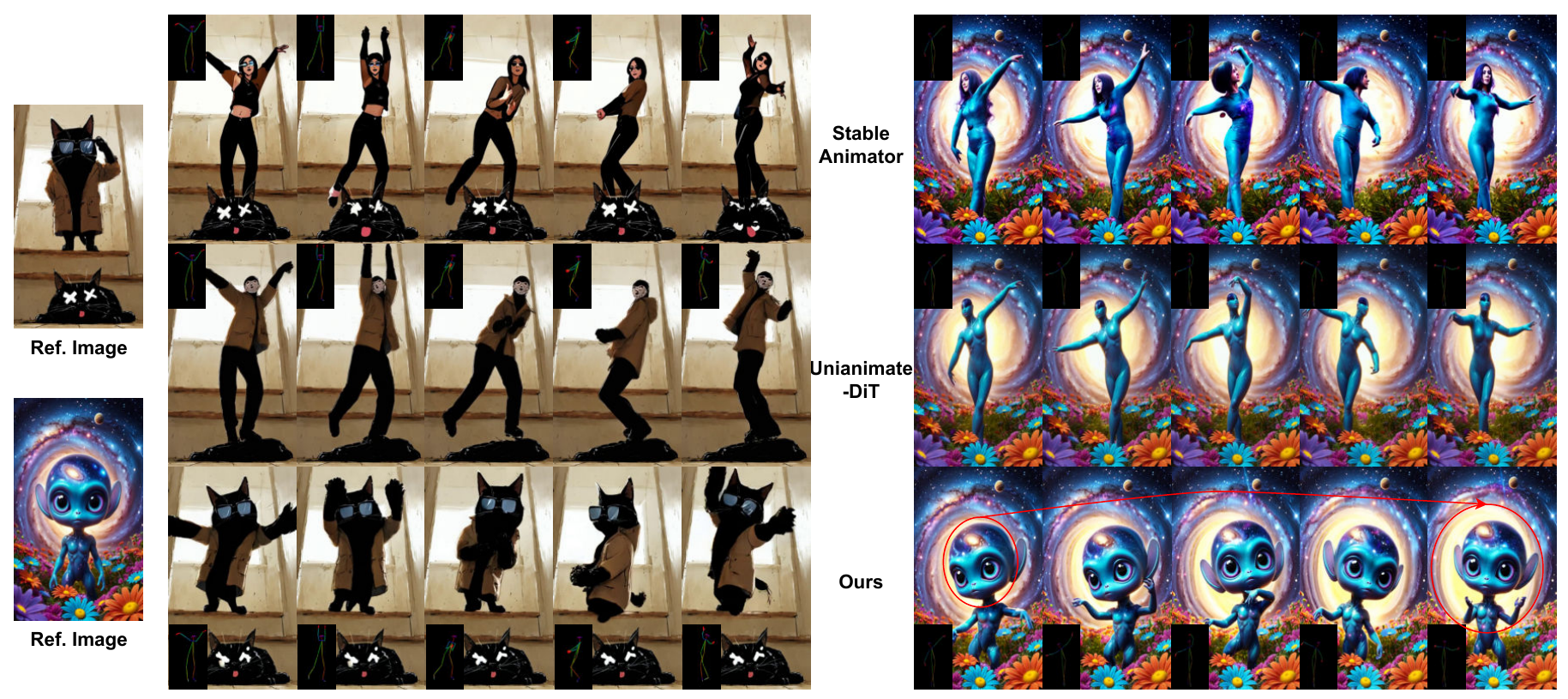}
  \captionsetup{justification=centering} 
    \caption{Comparison with SOTA methods.}
    \vspace{-5mm}
    \label{fig:more_comparison}
\end{figure*}

Following previous works which evaluate quantitative results in self-driven and reconstruction manner, we additionally compare our method with (a) GAN-based image animate works: FOMM~\cite{FOMM}, MRAA~\cite{MRAA}, LIA~\cite{wang2022latent}. (b) Diffusion model-based image animate works: DreamPose~\cite{karras2023dreampose}, MagicAnimate~\cite{magicanimate} and present the results in Tab.~\ref{tab:quantitative_cartoon_self_setting}, which indicates that our method achieves the best performance across all the metrics. Moreover, we provide the quantitative results on the human dataset (TikTok and Fashion) in Tab.~\ref{tab:quantitative_TikTok} and Tab.~\ref{tab:quantitative_fashion}, respectively. \method exceeds other SOTA methods, which demonstrates the superiority of \method on \textbf{both} anthropomorphic and human benchmarks. We notice that Unianimate-DiT achieves a comparable score to our method on human datasets. This is expected, as our Implicit Pose Indicator (IPI) is specifically designed to animate anthropomorphic characters. Unianimate-DiT performs well on human animation because it relies on carefully designed pose skeletons that align perfectly with human joints, but this approach falls short for anthropomorphic characters whose skeletons differ significantly from humans. For such characters, pose skeletons alone cannot provide sufficient driving guidance, as they lack the nuanced motion-related details found only in the original driving video. Consequently, as shown in Table~\ref{tab:quantitative_cartoon_cross_setting}, our method significantly outperforms Unianimate-DiT on the \benchmark.

\begin{figure}[t]
  \centering
  \includegraphics[width=\linewidth]{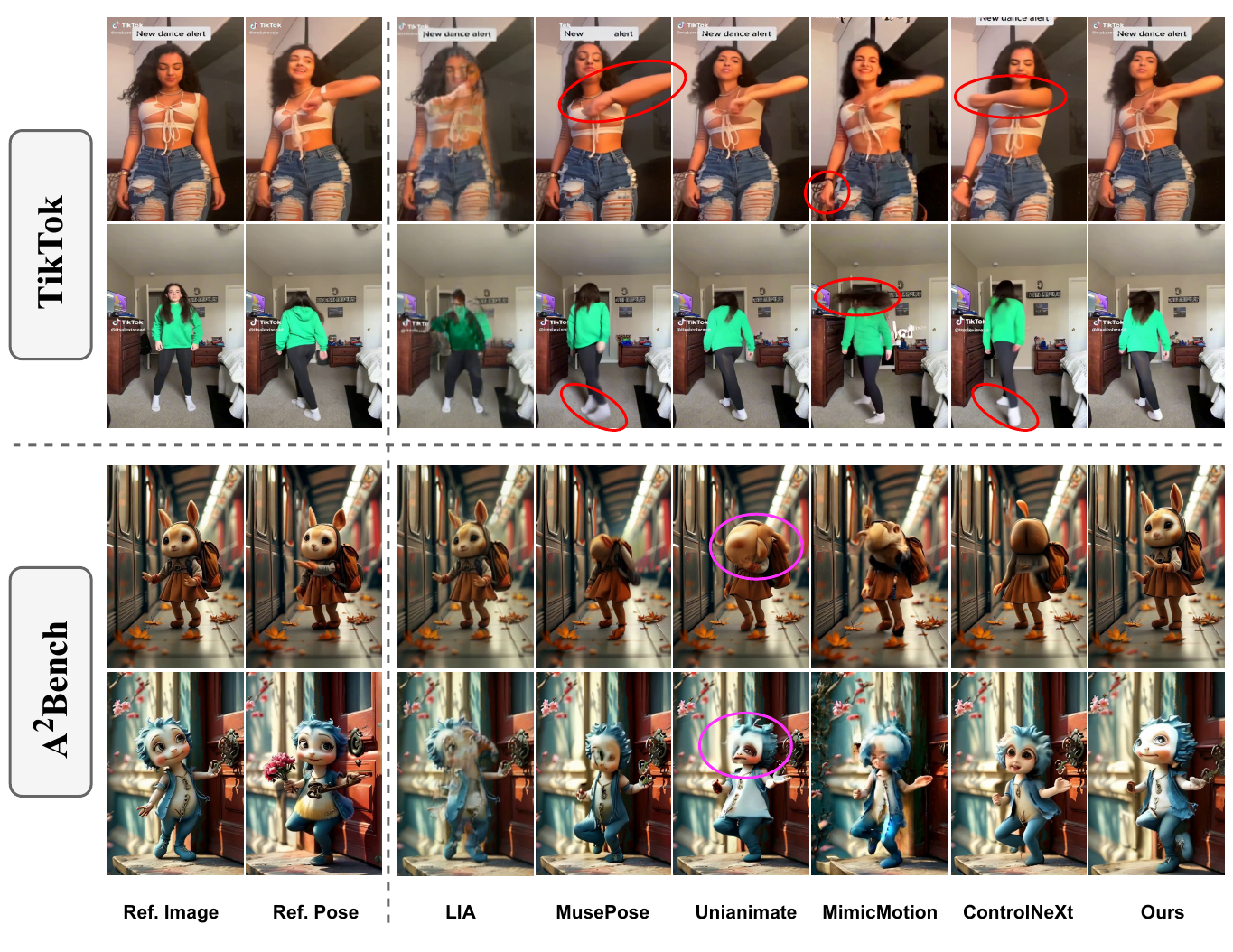}
  \captionsetup{justification=centering} 
    \caption{Visualization comparison on TikTok and \benchmark.}
    \label{fig:compare_tiktok_abench}
\end{figure}

\noindent \textbf{Qualitative Results.}
Qualitative comparisons of anthropomorphic animation are shown in Fig.~\ref{fig:compare}. We observe that GAN-based LIA~\cite{wang2022latent} does not generalize well, which can only work on a specific dataset like~\cite{siarohin2019first}. 
Benefiting from the powerful generative capabilities of the diffusion model, Animate Anyone~\cite{Animateanyone} renders a higher resolution image, but the identity of the image changes and do not generate an accurate reference pose motion. Although MusePose~\cite{musepose}, Unianimate~\cite{wang2024unianimate} and MimicMotion~\cite{mimicmotion2024} improve the accuracy of the motion transfer, these methods generate a unseen person, which is not the desired result. 
ControlNeXt combines the advantages of the above two types of methods, so maintains the consistency of identity and motion transfer to some extent, yet the results are somewhat unnatural and unsatisfactory, \textit{e.g.}, the ears of the rabbit and the legs of the banana in Fig.~\ref{fig:compare}.
In contrast, \method ensures both identity and consistency with the reference image while generating expressive and exaggerated figure motion, rather than simply adopting quasi-static motion of the target character.
The comparison with StableAnimator and Unianimate-DiT is displayed in Fig.~\ref{fig:more_comparison}, where our method not only successfully animates the cartoon characters while preserving their identities, but also effectively generates dynamic backgrounds.
Further, we present some long video comparisons in Fig.~\ref{fig:compare_2}. 
Unianimate generates a woman out of thin air who dances according to the given pose images. \method animates the reference image in a cute way while preserving appearance and temporal continuity, and it does not generate parts that do not originally exist. In summary, \method excels in maintaining appearance and producing precise, vivid animations with a high temporal consistency.

Considering that the other methods are primarily self-driven and trained on human characters, making them more suitable for inference in such settings, we additionally provide comparison results under a self-reconstruction setting on Tiktok and Abench. As shown in Fig.~\ref{fig:compare_tiktok_abench}, when there is a appreciably difference between the reference pose and the reference image, the GAN-based LIA~\cite{wang2022latent} produces noticeable artifacts. Thanks to the powerful generative capabilities of diffusion models, diffusion-based models generate higher-quality results. However, MusePose~\cite{musepose} and MimicMotion~\cite{mimicmotion2024} generate awkward arms and blurry hands, respectively, while ControlNeXt~\cite{peng2024controlnext} synthesizes incorrect movements. Only Unianimate~\cite{wang2024unianimate} can obtain results comparable to ours. Yet, when the reference image is a non-human character, even in a self-driven setting with the same training strategy as Unianimate, their results still show distorted heads. 
In contrast, our method consistently generates satisfactory results for both human and anthropomorphic characters, demonstrating its ability to drive $\texttt{X}$ character and highlighting its strong generalization.


\begin{table}[!t]
\caption{
Quantitative comparisons on Fashion dataset.
}
\renewcommand{\arraystretch}{1.15}
\setlength\tabcolsep{8.5pt}
\centering
\resizebox{\linewidth}{!}{
\begin{tabular}{l|cccc|c}
\shline
Method          & PSNR $\uparrow$ & PSNR* $\uparrow$ & SSIM $\uparrow$ & LPIPS $\downarrow$  & FVD $\downarrow$ \\ \shline

MRAA~\cite{MRAA}   &   -  &   - & 0.749 & 0.212 & 253.6   \\
TPS~\cite{TPS}   & - &   - & 0.746 & 0.213 & 247.5    \\
DPTN~\cite{DPTN}  & - &   24.00 & 0.907 & 0.060 & 215.1    \\
NTED~\cite{NTED}   & - &   22.03 & 0.890 & 0.073 & 278.9    \\
PIDM~\cite{bhunia2023person}  & - &   - & 0.713 & 0.288 & 1197.4    \\
DBMM~\cite{yu2023bidirectionally}    & -  &   {24.07} & 0.918 & 0.048 & 168.3    \\
\midrule
DreamPose~\cite{karras2023dreampose}  &  -  &   - & 0.885 & 0.068  & 238.7        \\
DreamPose \emph{w/o Finetune}~\cite{karras2023dreampose}  &  34.75 &   -  & 0.879 & 0.111 &  279.6        \\
Animate Anyone~\cite{Animateanyone}  &  {{38.49}} &   - & {0.931} & {0.044} & {81.6}  
\\

Unianimate~\cite{wang2024unianimate}  & {37.92}          &   {27.56}        & {0.940}            & {0.031}                                            & \textbf{68.1}         \\

MimicMotion~\cite{mimicmotion2024}   & -          &   27.06       & 0.928            & 0.036                                           & 118.48        \\
StableAnimator~\cite{tu2025stableanimator}   &  {37.15}          &   27.86       & 0.942           & 0.029                                          & 75.8       \\
Unianimate-DiT~\cite{tu2025stableanimator}    & {37.60}          &   \textbf{27.98}       & 0.945           & \textbf{0.027}                                          & 72.3          \\
\rowcolor{Gray}
\textbf{Ours-3D Unet~\cite{animatex}}  & {36.73}          &   {27.78}        & {0.940}            & {0.030}                                            & {79.4}         \\
\rowcolor{Gray}
\textbf{Ours-DiT(\method)}  &  \textbf{38.71}         &   27.95       & \textbf{0.947}           & 0.028                                           & {70.9}      \\

\shline
\end{tabular} 
}

\label{tab:quantitative_fashion}
\vspace{-4mm}
\end{table}

\begin{figure}
  \includegraphics[width=0.9\linewidth]{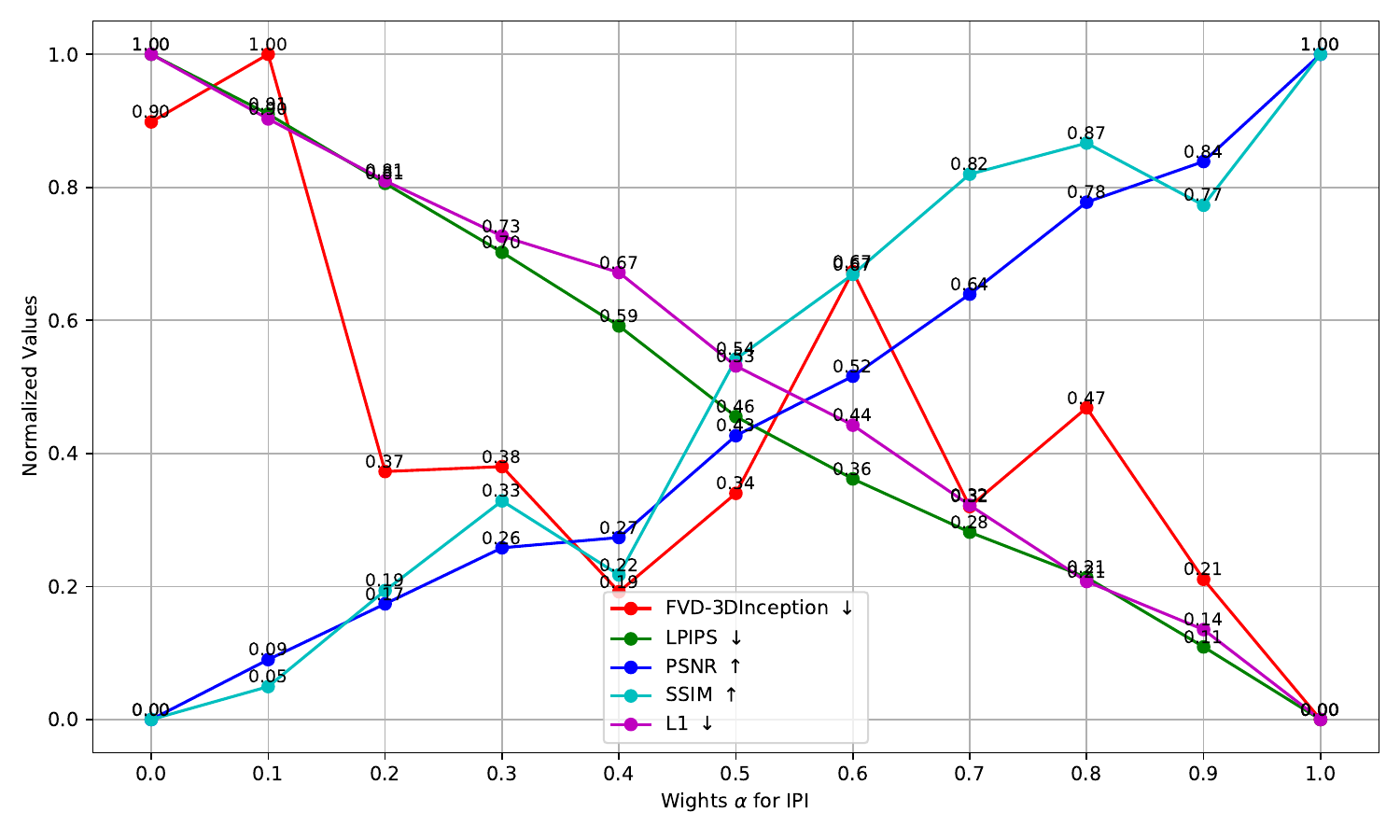}
  \captionsetup{justification=centering} 
    \caption{Ablation study on the weight $\alpha$ of IPI.}
    \label{fig:qformer_weight}
\end{figure}

\begin{table*}[!t]
\caption{
User study results.
}
\renewcommand{\arraystretch}{1}
\setlength\tabcolsep{6.5pt}
\centering
\resizebox{\linewidth}{!}{
\begin{tabular}{l|cccccc} 
\shline
Method         &  Moore-AA & MimicMotion     & ControlNeXt & {MusePose} & Unianimate      & \textbf{\method}  \\ \shline

Identity preservation $\uparrow$ & {60.4\%}     & {14.8\%} & 52.0\%      & 31.3\%            & {43.0\%} & \textbf{98.5\%} \\
Temporal consistency $\uparrow$ & {19.8\%}     & 24.9\%          & 36.9\%      & 43.9\%            & 81.1\%          & \textbf{93.4\%} \\
Visual quality  $\uparrow$      & {27.0\%}     & 17.2\%          & 40.4\%      & 40.3\%            & 79.3\%          & \textbf{95.8\%}        \\

\shline
\end{tabular} 
}
\vspace{-4mm}

\label{tab:user_study}
\end{table*}

\noindent \textbf{User Study.} 
To estimate the quality of our method and SOTAs from human perspectives, we conduct a blind user study with 10 participants. Specifically, we randomly select 10 characters from \benchmark and collect 10 driving video from the website. For each of 6 methods tested, 10 animation clips are generated, resulting in a total of 60 clips. Each participant is presented two results generated by different methods for the same set of inputs and asked to choose which one is better in terms of \textit{visual quality}, \textit{identity preservation}, and \textit{temporal consistency}. This process is repeated $C^6_2$ times. The results are summarized in Tab.~\ref{tab:user_study}, where our method noticeably outperforms other methods in all aspects.

\begin{table*}[t]

	\caption{Quantitative results for different probabilities of using pose transformation.}
	\centering
	\resizebox{0.75\linewidth}{!}{
	\begin{tabular}{l|cccc|cccc}
		\toprule
		\multicolumn{1}{c}{\multirow{2}[4]{*}{\textbf{Method}}} & \multicolumn{4}{c}{\benchmark} & \multicolumn{4}{c}{\textbf{TikTok~\cite{TikTokdata}}}\\
		\cmidrule(lr){2-5}  \cmidrule(lr){6-9}  \multicolumn{1}{c}{} & \multicolumn{1}{c}{SSIM$\uparrow$} & \multicolumn{1}{c}{FID$\downarrow$}  & \multicolumn{1}{c}{FID-VID$\downarrow$}& \multicolumn{1}{c}{FVD$\downarrow$}& \multicolumn{1}{c}{SSIM$\uparrow$} & \multicolumn{1}{c}{FID$\downarrow$}  & \multicolumn{1}{c}{FID-VID$\downarrow$}& \multicolumn{1}{c}{FVD$\downarrow$}
		\\

		\midrule
\textbf{100\%}  & \textbf{0.452}       & \textbf{26.11} & \textbf{32.23}       & \textbf{703.87}       & 0.802          & 55.26          & 17.47          & 138.36          \\
\textbf{98\%}  & { {0.448}} & {{26.93} }    & {{37.67} } & {{775.24}} & 0.797          & 55.81          & 16.28          & {{129.48} }    \\
\textbf{95\%}  & 0.447                & 27.46          & 39.21       & 785.55                & {{0.804} }    & \textbf{52.72} & {{14.61} }    & \textbf{124.92} \\
\textbf{90\%} & 0.444                & 27.15          & 38.03       & 775.38                & \textbf{0.806} & {{52.81} }    & 14.82          & 139.01          \\
\textbf{80\%} & 0.442                & 29.13          & 47.93       & 803.97                & 0.802          & 54.51          & \textbf{14.42} & 133.78       \\

		\bottomrule
	
	\end{tabular}%
	}

	\label{tab:training_ratio}%
    \vspace{-4mm}
\end{table*}

\begin{figure}[t]
  \centering
  \includegraphics[width=\linewidth]{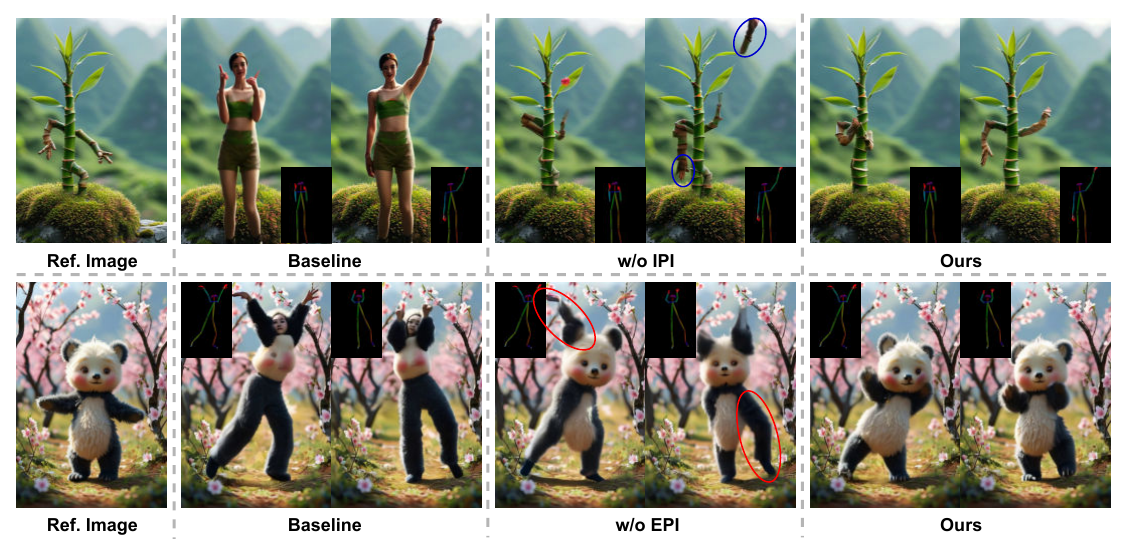}
  \captionsetup{justification=centering} 
    \caption{Visualization of ablation study on IPI and EPI.}
    \vspace{-4mm}
    \label{fig:ablation}
\end{figure}

\subsection{Ablation Study}

\noindent \textbf{Ablation on Implicit Pose Indicator.} Given the computational overhead of the DiT architecture for both training and inference compared to a 3D Unet, we adopt a 3D Unet as our backbone for conducting the ablation studies on IPI and EPI, which allows for a more efficient validation of these components.
To analyze the contributions of Implicit Pose Indicator, we remove it from \method as w/o IPI and compare it with Baseline and \method. From the first row of Fig.~\ref{fig:ablation}, we observe that Baseline generates a person whose appearance is appreciably distinct from the reference image. With the help of EPI, this problem is mildly mitigated. However, due to the absence of IPI, compared to Ours, there are still strange things and human-like hands appearing, as indicated by the blue circle. For more detailed analysis about the structure of IPI, we set up several variants: (1) remove IPI: w/o IPI. (2) remove learnable query: w/o LQ. (3) remove DWPose query: w/o DQ. (4) set IPI and spatial {A}ttention to {P}arallel: {PA}. (5) set CLIP features as {Q} and DWPose as {K,V} in IPI: {KV\_Q}. The quantitative results are shown in Tab.~\ref{tab:ablation_study}. It can be seen that removing the entire {IPI} presents the worst performance. By modifying the IPI module, although it improves on the {w/o IPI}, it still falls short of the final result of \method, which suggests that our current IPI structure achieves the best performance.

Since {IPI} is embedded in \method in the form of residual connection, i.e., $x = x + \alpha IPI(x)$, we also explore the impact of the weight $\alpha$ of {IPI} on performance as illustrated in Fig.~\ref{fig:qformer_weight}, as $\alpha$ increases from 0 to 1, all metrics show a stable improvement despite some fluctuations. The best performance is achieved when $\alpha$ is set to 1, so we empirically set $\alpha$ to 1.

\begin{table*}[!t]

\caption{
Quantitative comparisons with existing methods on TikTok dataset.}

\renewcommand{\arraystretch}{1}
\setlength\tabcolsep{7pt}
\centering
\resizebox{\linewidth}{!}{
\begin{tabular}{l|ccccc|c}
\shline
Method          & L1 $\downarrow$ & PSNR $\uparrow$ & PSNR* $\uparrow$ & SSIM $\uparrow$ & LPIPS $\downarrow$  & FVD $\downarrow$ \\ \shline
FOMM~\cite{FOMM} $_{\color{gray}{\text{(NeurIPS19)}}}$   & 3.61E-04        & -     & 17.26       & 0.648           & 0.335                                       & 405.22           \\
MRAA~\cite{MRAA} $_{\color{gray}{\text{(CVPR21)}}}$   & 3.21E-04        & -     & 18.14       & 0.672           & 0.296                                       & 284.82           \\
TPS~\cite{TPS} $_{\color{gray}{\text{(CVPR22)}}}$   & 3.23E-04        & -     & {18.32}       & 0.673           & 0.299                                       & 306.17           \\
\midrule
DreamPose~\cite{karras2023dreampose} $_{\color{gray}{\text{(ICCV23)}}}$  & 6.88E-04        & 28.11    & 12.82       & 0.511           & 0.442                            & 551.02           \\
DisCo~\cite{disco} $_{\color{gray}{\text{(CVPR24)}}}$            & 3.78E-04            & 29.03      & 16.55       & 0.668             & 0.292                                      & 292.80              \\
MagicAnimate~\cite{magicanimate} $_{\color{gray}{\text{(CVPR24)}}}$   & 3.13E-04    & 29.16  & -         & 0.714           & 0.239                                 & 179.07           \\
Animate Anyone~\cite{Animateanyone} $_{\color{gray}{\text{(CVPR24)}}}$   & -            & 29.56      & -       & 0.718             & 0.285                                            & 171.90              \\
Champ~\cite{zhu2024champ} $_{\color{gray}{\text{(ECCV24)}}}$ 
& {2.94E-04}            & {29.91}        & -     & {0.802}            & {0.234}                                        & {160.82}         \\

Unianimate~\cite{wang2024unianimate} $_{\color{gray}{\text{(SCIS25)}}}$  & {2.66E-04}       & {30.77}    & {20.58}             & {0.811}            & {0.231}                                          &{148.06}         \\

MusePose~\cite{musepose} $_{\color{gray}{\text{(ArXiv24)}}}$  & 3.86E-04       & -    & 17.67            & 0.744            & 0.297                               &215.72         \\

MimicMotion~\cite{mimicmotion2024} $_{\color{gray}{\text{(ICML25)}}}$  & 5.85E-04       & -    & 14.44             & 0.601            & 0.414                                          &232.95         \\

ControlNeXt~\cite{peng2024controlnext} $_{\color{gray}{\text{(ArXiv24)}}}$  & 6.20E-04       & -    & 13.83             & 0.615           & 0.416                                         &326.57        \\

StableAnimator~\cite{tu2025stableanimator} $_{\color{gray}{\text{(CVPR25)}}}$  &2.68E-04    & 30.85    & 20.86            & 0.809           & 0.230 &137.55     \\

Unianimate-DiT~\cite{tu2025stableanimator} $_{\color{gray}{\text{(CVPR25)}}}$  & 2.65E-04       & 30.95    & \textbf{20.98}             & 0.815           & 0.229                                        &136.10        \\

\rowcolor{Gray}

\textbf{Ours-3D Unet~\cite{animatex} $_{\color{gray}{\text{(ICLR25)}}}$}  & {2.70E-04}       & {30.78}    & {20.77}             & {0.806}            & {0.232}  &{139.01}         \\

\rowcolor{Gray}
\textbf{Ours-DiT (\method)}  &  \textbf{2.64E-04}       & \textbf{31.05}    & {20.95}             & \textbf{0.817}            & \textbf{0.228}  &\textbf{135.20}         \\

\shline
\end{tabular} 
}

%
\label{tab:quantitative_TikTok}
\vspace{-2mm}
\end{table*}

\begin{figure}[t]
  \centering
  \includegraphics[width=1\linewidth]{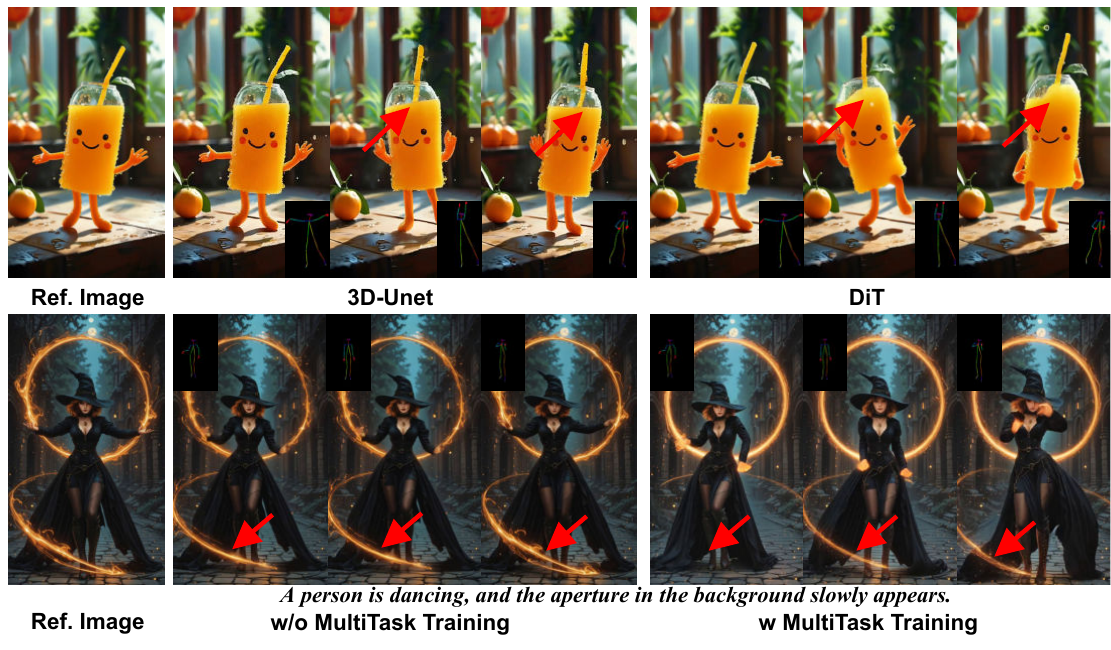}
  \captionsetup{justification=centering} 
    \caption{Ablation study on backbone and training strategy.}
    \label{fig:new_ablation}
\vspace{-4mm}
\end{figure}

\begin{table*}[!t]
\caption{
Quantitative results of ablation study.
}
\renewcommand{\arraystretch}{1}
\setlength\tabcolsep{12pt}
\centering
\resizebox{\linewidth}{!}{
\begin{tabular}{l|ccccc|cc}
\shline
Method         & PSNR* $\uparrow$ & SSIM $\uparrow$ & L1 $\uparrow$ & LPIPS $\downarrow$ & FID $\downarrow$  &FID-VID $\downarrow$  & FVD $\downarrow$ \\ \shline
w/o IPI            & 13.30          & 0.433          & 1.35E-04          & {0.454} & 32.56          & 64.31 & 893.31 \\
w/o LQ           & {13.48} & 0.445          & 1.76E-04          & {0.454} & 28.24          & 42.74 & 754.37 \\
w/o DQ         & 13.39          & 0.445          & {1.01E-04}          & 0.456          & 30.33          & 62.34 & 913.33 \\
PA         & 13.25          & 0.436          & 1.11E-04          & 0.464          & 27.63          & 46.54 & 785.36 \\
KV\_Q           & 13.34          & 0.443          & 1.17E-04          & 0.459          & 26.75          & 42.14 & 785.69 \\
\midrule

w/o EPI         & 12.63          & 0.403          & 1.80E-04          & 0.509          &     42.17      & 58.17 &   948.25\\

w/o Add               & 13.28          & 0.442          & 1.56E-04          & 0.459          & 34.24          & 52.94 & 804.37 \\
w/o Drop              & 13.36          & 0.441          & {1.94E-04} & 0.458          & {26.65}          & 44.55 & 764.52 \\
w/o BS              & 13.27          & 0.443          & {1.08E-04} & 0.461          & 29.60          & 56.56 & 850.17 \\
w/o NF              & 13.41          & {0.446}          & 1.82E-04          & 0.455          & 29.21          & 56.48 & 878.11 \\
w/o AL          & 13.04          & 0.429          & 1.04E-04          & 0.474          & 27.17          & {33.97} & 765.69 \\
w/o Rescalings & 13.23          & 0.438          & 1.21E-04          & 0.464          & 27.64          & 35.95 & {721.11} \\
w/o Realign     & 12.27          & 0.433          & 1.17E-04          &    0.434       &  34.60         & 49.33 &860.25 \\

\midrule

{3D Unet}  & {13.60} & {0.452} &{ 1.02E-04} & {0.430} & {26.11} & {32.23} & {703.87}   \\

{DiT}  &\textbf{13.98} & {0.460} & \textbf{9.28E-05} & {0.423} & {25.35} & {31.20} & \textbf{678.95}  \\

\textbf{\method}  & {13.94} & \textbf{0.462} &{ 9.34E-05} & \textbf{0.421} & \textbf{25.14} & \textbf{31.09} & {681.42}   \\

\shline
\end{tabular} 
}
\vspace{-4mm}

\label{tab:ablation_study}
\end{table*}

\noindent \textbf{Ablation on Explicit Pose Indicator.} We demonstrate the visual results of ablating EPI setting in the second row of Fig.~\ref{fig:ablation} by removing EPI. Without EPI, although the appearance of the panda is preserved thanks to IPI, the model incorrectly treats the panda's ears as arms and forcibly stretches the legs to match the length of the legs in the pose image indicated by red circles. In contrast, these issues are completely resolved by the assistance of EPI. We further conduct more detailed ablation experiments for different pairs of pose transformations by (1) removing the entire EPI: w/o EPI. (2)\&(3) removing adding and dropping parts; canceling the change of the length of (4) body and should: {w/o BS}; (5) neck and face: w/o NF; (6) arm and leg: {w/o AL}; (7) removing all rescaling process: {w/o Rescalings}; (8) remove another person pose alignment: {w/o Realign}. From the results displayed in Tab.~\ref{tab:ablation_study}, we find that each pose transformation contributes compared to w/o EPI, with aligned transformations with another person's pose contributing the most. It suggests that maintaining the overall integrity of the pose while allowing for some variations is the most important factor, and EPI also learns the overall integrity of the pose. The final result indicates that all the transformations together achieve the best performance.

To explore the effect of different probabilities $\lambda$ of using pose transformation for {EPI} on the model performance, we set $\lambda$ as 100\%, 98\%, 95\%, 90\% and 80\% for the ablation experiments on two datasets. The results presented in Tab.~\ref{tab:training_ratio} suggest that a high $\lambda$ performs better on \benchmark, i.e., it performs better when the reference image and pose image are not aligned, but harms performance on the TikTok dataset, i.e., when the reference image and pose image are strictly aligned. In contrast, a relatively low $\lambda$, e.g., 90\%, would be in this case perform better. It is reasonable that in the case of strict alignment, we expect the pose to provide a strictly accurate motion source, and thus need to reduce the percentage $\lambda$ of pose transformation. However, in the non-strictly aligned case, we expect the pose image to provide an approximate motion trend, so we need to increase $\lambda$.

Since the anchor poses are chosen from the entire training set, we further conduct the statistical analysis for rescaling ratio. First, we randomly sample a driven pose $I^p$ and then traverse the entire pose pool, treating each pose in the pool as an anchor pose to calculate the rescaling ratio. We repeat this process 10 times. Finally, we divide the range from 0.001 to 10 into 10 intervals, counting the proportion of rescaling ratios that fell within each interval. We analyze the proportions of other important parts like shoulder length, body length, upper/lower arm length, upper/lower leg length. As shown in Tab.~\ref{tab:statistical_analysis}, the overall distribution covers a wide range (from 0.001 to 10.0), which allows the model to learn poses of various characters, encompassing non-human subjects.

\begin{table*}[!t]
\caption{Statistical analysis for rescaling ratio.}
\renewcommand{\arraystretch}{1.15}
\setlength\tabcolsep{6.5pt}
\centering
\resizebox{\linewidth}{!}{
\begin{tabular}{l|cccccc}
\shline
Interval         & Shoulder Length & Body Length & Upper Arm Length & Lower Arm Length & Upper Leg Length & Lower Leg Length \\ \shline
$[0.001, 0.1)$    & 0.19\%          & 0.14\%      & 0.05\%          & 0.08\%          & 0.05\%          & 0.81\%         \\
$[0.1, 0.3) $     & 1.52\%          & 5.73\%      & 4.04\%          & 3.22\%          & 0.59\%          & 4.60\%          \\
$[0.3, 0.5)$      & 12.21\%         & 18.57\%     & 15.28\%         & 7.63\%          & 4.26\%          & 5.65\%           \\
$[0.5, 0.7) $     & 15.33\%         & 16.93\%     & 12.97\%         & 7.54\%          & 12.02\%         & 9.61\%           \\
$[0.7, 1.0)  $    & 20.07\%         & 18.48\%     & 17.15\%         & 11.35\%         & 24.86\%         & 19.53\%          \\
$[1.0, 1.5) $     & 22.09\%         & 18.63\%     & 17.56\%         & 15.38\%         & 27.90\%         & 24.89\%          \\
$[1.5, 2)   $     & 10.07\%         & 8.34\%      & 7.93\%          & 11.73\%         & 14.31\%         & 14.47\%          \\
$[2.0, 3.0) $     & 9.75\%          & 6.52\%      & 7.73\%          & 16.19\%         & 11.83\%         & 15.28\%          \\
$[3.0, 6.0)  $    & 6.33\%          & 6.28\%      & 10.93\%         & 18.40\%         & 2.73\%          & 4.30\%           \\
$[6.0, 10.0) $    & 2.43\%          & 0.37\%      & 6.37\%          & 8.47\%          & 1.45\%          & 0.85\%           \\
\shline
\end{tabular} 
}

\label{tab:statistical_analysis}
\end{table*}

\noindent \textbf{Ablation on Backbone and Training Strategy.} We first conduct an ablation study on the model's backbone, comparing the commonly-used 3D Unet with the recently developed DiT framework. As illustrated in Fig.~\ref{fig:new_ablation}, the videos generated by the DiT backbone reveal a higher degree of fidelity and continuity, demonstrating more lifelike movements compared to those from the 3D Unet. Notably, the DiT-based model exhibits a superior ability to perceive and render character-specific dynamics. For instance, with the cartoon juice character shown in Fig.~\ref{fig:new_ablation}, the DiT-based result shows the liquid inside the cup swaying in response to the character's movement that the 3D UNet model fails to capture. Building on the DiT backbone, we then ablate our proposed multi-task training strategy to validate its contribution. As shown in Fig.~\ref{fig:new_ablation}, without this strategy, the light aperture in the video remains static. In contrast, our full method, incorporating multi-task training, successfully renders the dynamic emergence of the aperture, making the result more compelling and visually appealing. The quantitative results in Tab.~\ref{tab:ablation_study} further corroborate the significant contributions of both the DiT architecture and the multi-task training strategy.

In summary, we can draw conclusions: (1) IPI facilitates the preservation of appearance and prevents the generation of content that does not exist in the reference image like human arms. (2) EPI prevents the forced alignment of a pose image that is not naturally aligned with the reference image during animation, thus avoiding the unintended animation of parts that should remain static like the panda's ears shown in Fig.~\ref{fig:ablation}. (3) The adoption of the DiT architecture, in place of a 3D Unet, yields superior video rendering quality and enhances temporal smoothness between frames. (4) The multi-task training strategy enables the model to generate dynamic backgrounds, significantly boosting the naturalness and realism of the final results.

\begin{figure}[t]
  \centering
  \includegraphics[width=0.95\linewidth]{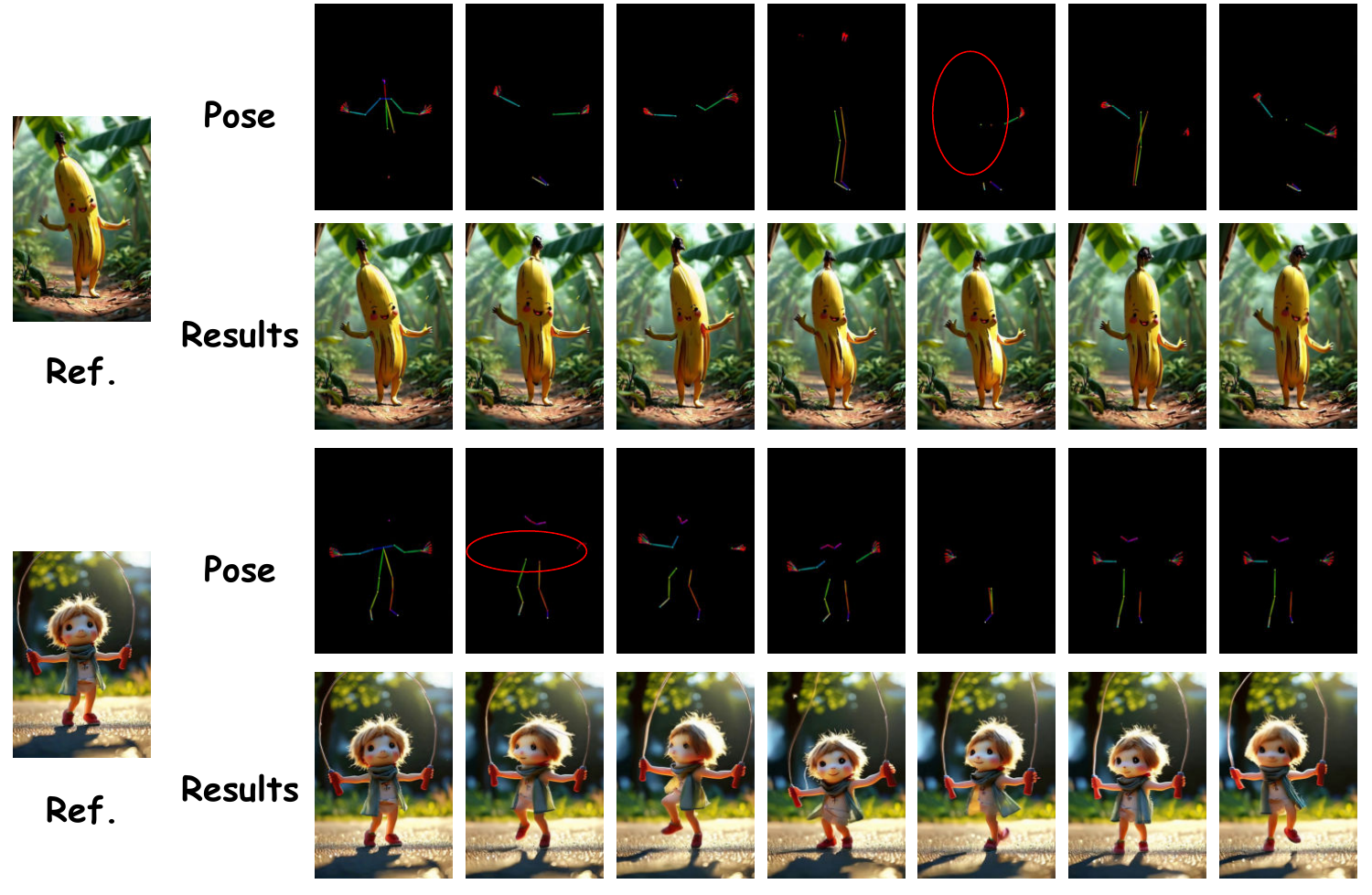}
  \captionsetup{justification=centering} 
    \caption{Visualization of the robustness of \method.}
    \label{fig:robustness}
    \vspace{-4mm}
\end{figure}

\subsection{More Interesting Results}
\noindent \textbf{Robustness.}
Our method demonstrates robustness to both input \texttt{X} character and pose variations. On the one hand, as shown in Fig.~\ref{fig:teaser}, our approach successfully handles inputs from diverse subjects, including characters vastly different from humans, such as those without limbs. Despite these variations, our method consistently produces satisfactory results, showcasing its robustness to the input reference images. On the other hand, as illustrated in Fig.~\ref{fig:robustness}, even when the pose images exhibit body part omissions, our method correctly interprets the intended motion. This highlights the robustness of \method to different poses.

\begin{figure}[t]
  \centering
  \includegraphics[width=1\linewidth]{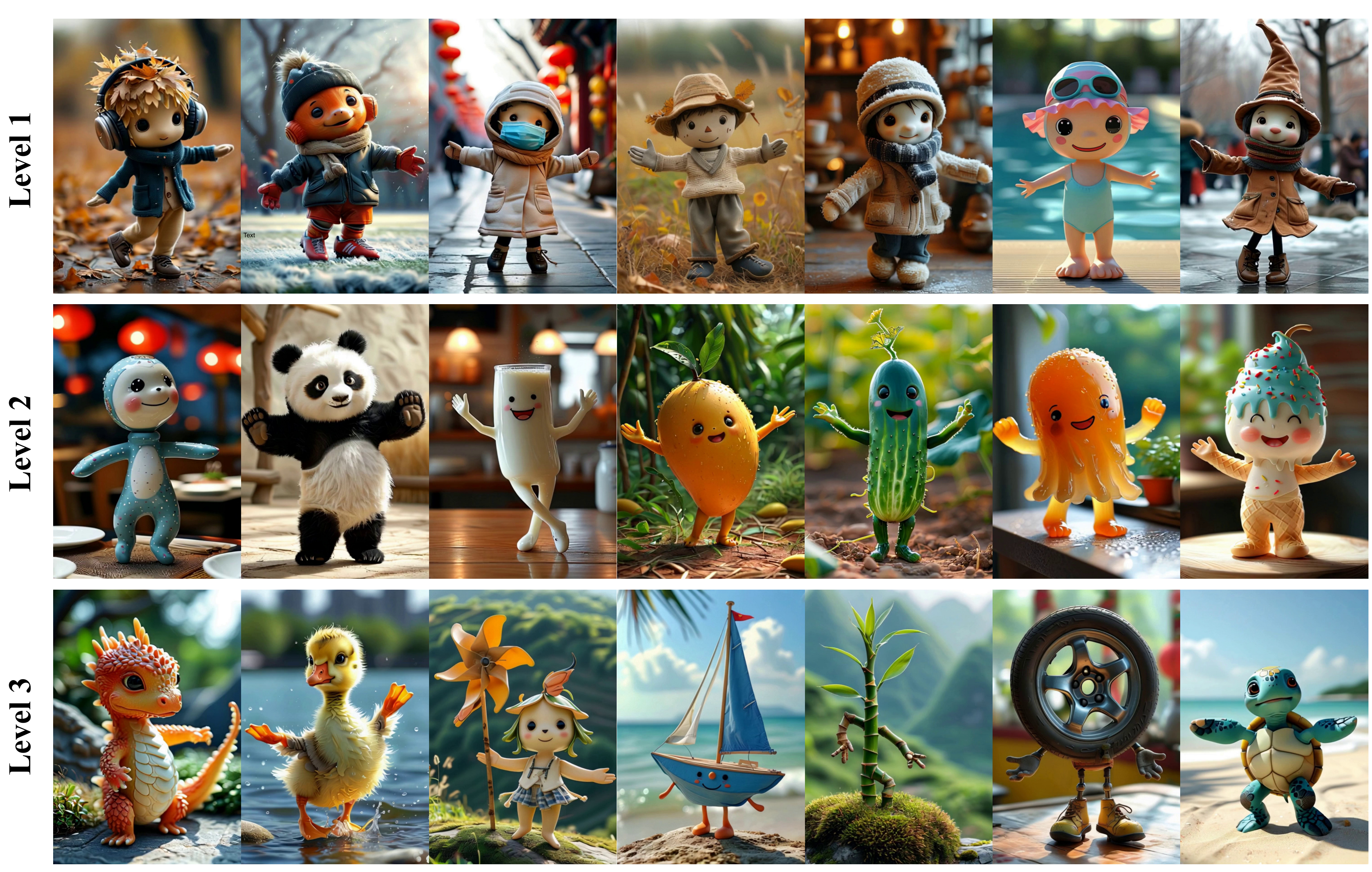}
  \captionsetup{justification=centering}
    \caption{Difficulty levels in \benchmark.}
    \label{fig:difficulties}
\vspace{-6mm}
\end{figure}

\begin{table*}[t]
\caption{
Results on \benchmark of different difficult levels.
}
\renewcommand{\arraystretch}{1}
\setlength\tabcolsep{6.5pt}
\centering
\resizebox{\linewidth}{!}{
\begin{tabular}{l|ccccccc} 
\shline
Model-Level   & PSNR* $\uparrow$ & SSIM $\uparrow$ & L1 $\downarrow$       & LPIPS  $\downarrow$ & FID  $\downarrow$   & FID-VID  $\downarrow$ & FVD $\downarrow$    \\ \shline

\method-level1    & 13.96   & 0.461  & 9.67E-05   & 0.418   & 24.24   & 31.37     & 681.53   \\
\method-level2    & 13.74   & 0.457  & 9.82E-05   & 0.429   & 26.12   & 32.19     & 693.63   \\
\method-level3    & 13.17   & 0.442  & 1.11E-04   & 0.437   & 27.34   & 35.64     & 721.41   \\

\shline
\end{tabular} 
}

\label{tab:abench_level}
\end{table*}

\noindent \textbf{Difficulty Level of \benchmark.} We add the difficulty level split for \method. As shown in Figure ~\ref{fig:difficulties}, we categorize the videos in \benchmark into three difficulty levels: Level 1, Level 2, and Level 3. The classification is based on their appearance characteristics. {First}, we classify characters that have body shapes and other appearance features similar to humans, as shown in the first row of Fig.~\ref{fig:difficulties}, into the easiest, Level 1 category. These characters are generally simpler to drive, produce fewer artifacts, and have better motion consistency. {In contrast}, characters that maintain more distinct structural features from humans, such as dragons and ducks in the third row of Fig.~\ref{fig:difficulties}, are classified into the most difficult Level 3 category. These characters often preserve their original structures (\textit{e.g.}, a duck's webbed feet and wings), which makes balancing identity preservation and motion consistency more challenging. 
To ensure identity preservation, the consistency of motion may be compromised, and vice versa. Additionally, images involving interactions between characters, objects, environments, and backgrounds are also placed in Level 3, as they increase the difficulty for the model to distinguish the parts that need to be driven from those that do not. 
{Videos in between these two categories}, like those in the second row of Fig.~\ref{fig:difficulties}, are classified as Level 2. These characters often strike a good balance between anthropomorphism and their original form, making them easier to animate with better motion consistency than Level 3 characters and more interesting results than Level 1 characters. We evaluate the results of \method and UniAnimate for each subset. In this experiment, we adopt a 3D Unet as our backbone for efficient validation. As shown in Tab.~\ref{tab:abench_level}, as the difficulty increases, each evaluation result shows a decline.

\begin{figure*}[t]
  \centering
  \includegraphics[width=\linewidth]{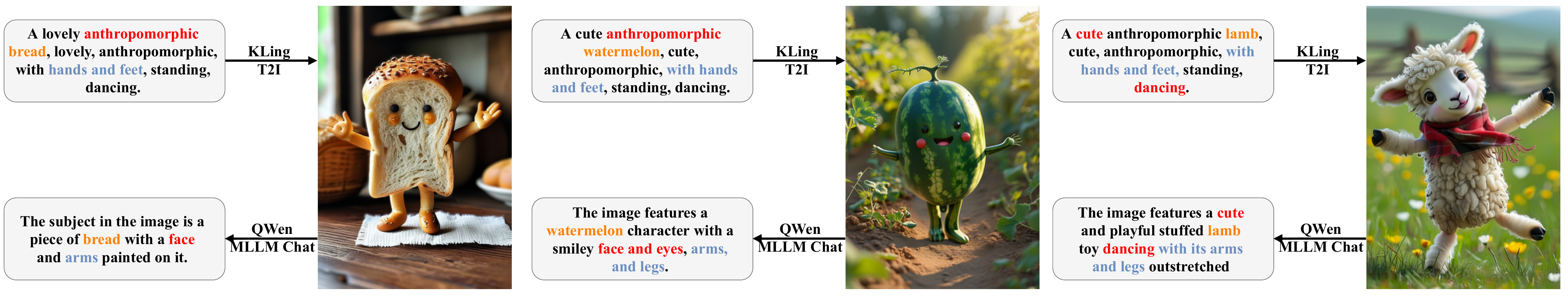}
  \captionsetup{justification=centering} 
    \caption{Prompts, generated images by T2I in \benchmark, and logical answers from QWen.}
    \label{fig:abench_qwen}
    \vspace{-4mm}
\end{figure*}

\section{Discussion}
\label{sec:discussion}

\noindent \textbf{Motivation of T2I+I2V for \benchmark.} The choice to use T2I models stems from a clear need: current T2V models often struggle with imaginative and logically complex inputs, such as "\textit{personified refrigerators}" or "\textit{human-like bees}". T2I models offer strict logic and imagination in these scenarios, allowing to generate reasonable cartoon characters as the ground-truth. To prove this point, as shown in Tab.~\ref{tab:benchmark_table}, we assess the semantic accuracy of \benchmark using CLIP scores, which are commonly used to evaluate whether the semantic logic of images and text is strictly aligned (\textit{i.e.}, Does the generated ``\textit{human-like bee}'' maintain the visual essence of a bee while seamlessly incorporating human-like features, such as hands and feet?). We also add other metrics from VBench~\cite{vbench}, such as \textit{Background Consistency}, \textit{Motion Smoothness}, \textit{Aesthetic Quality} and \textit{Image Quality}, to assess the spatial and temporal consistency of the videos in \benchmark.  For comparison, we also evaluate the publicly available TikTok and Fashion datasets using the same metric. As shown in Tab.~\ref{tab:benchmark_table}, \benchmark achieves the highest level of strict logical alignment. \benchmark outperforms TikTok in all aspects and achieve comparable scores to Fashion, where both datasets are collected from real-world scenarios. It demonstrates that the video generated by our method has the same level of spatial and temporal consistency as the real videos.

Furthermore, we input the images from \benchmark into a multimodal large language model (MLLM) QWen~\cite{bai2023qwen} with logical reasoning, to conduct a logical analysis of the visual outputs generated by the T2I model. The results, shown in Fig.~\ref{fig:abench_qwen}, reveal that the image descriptions answered by the MLLM closely aligns with our input prompts, which verifies again that the data in \benchmark maintains strict logic.

\begin{table}[t]
\caption{%
   Quantitative results of different benchmarks.
}
\setlength\tabcolsep{2pt}
\def\w{10pt} 

\centering\footnotesize
\begin{tabular}{l@{\extracolsep{10pt}}c@{\extracolsep{10pt}}c@{\extracolsep{10pt}}c@{\extracolsep{10pt}}c@{\extracolsep{10pt}}c@{\extracolsep{10pt}}c@{\extracolsep{10pt}}c@{\extracolsep{10pt}}c}
\shline
\multirow{2}{*}{\textbf{Benchmark}}             & \textbf{CLIP}   & \textbf{Background} &    \textbf{Motion}  & \textbf{Aesthetic} & \textbf{Image}  \\\vspace{-4mm}\\
 & \textbf{Score} & \textbf{Consistency} & \textbf{Smoothness} & \textbf{Quality}& \textbf{Quality} \\
    \hline
TikTok & {26.92} & 94.10 \%                  & 99.05 \%                     & {55.14} \%   & {62.54} \%           \\
Fashion    & {20.18} & \textbf{98.25}    \%              & \textbf{99.45 }      \%               & {49.62} \% & 49.96  \%      \\
\hline
\benchmark & \textbf{33.24} & {96.66}   \%               & {99.39}           \%           & \textbf{69.86}  \%          & \textbf{69.32}   \%      \\
\shline
\end{tabular}

  \label{tab:benchmark_table}%
  \vspace{-4mm}
\end{table}%

\noindent \textbf{Limitation and Future Work}
Although our method has made remarkable progress, it still has certain limitations. Firstly, its ability to model hands and faces remains insufficient, a limitation commonly faced by most current generative models. While our \textbf{IPI} leverages CLIP features to extract implicit information such as motion patterns from the driving video, mitigating the reliance on potentially inaccurate hand and face detection by DWPose, there is still a gap between our results and the desired realism. Secondly, due to the multiple denoising steps in the diffusion process, even though we replace the transformer with a more efficient Mamba model for temporal modeling, \method still cannot achieve real-time animation. In future work, we aim to address these two limitations. Additionally, we will focus on studying interactions between the character and the surrounding environment, such as the background, as a key task to resolve.

\noindent \textbf{Ethical Considerations}
Our approach focuses on generating high-quality character animation videos, which can be applied in diverse fields such as gaming, virtual reality, and cinematic production. By providing body movement, our method enables animators to create more lifelike and dynamic characters. However, the potential misuse of this technology, particularly in creating misleading or harmful content on digital platforms, is a concern. While greatly progress has been made in detecting manipulated animations~\cite{boulkenafet2015face, wang2020deep, yu2020searching}, challenges remain in accurately identifying sophisticated forgeries. We believe that our animation results can contribute to the development of better detection techniques, ensuring the responsible use of animation technology across different domains.

\section{Conclusions}

\label{sec: conclusion}
In this study, we present \method, a novel approach to character animation capable of generalizing across different types of characters named \texttt{X} with dynamic backgrounds. To address the imbalance between identity preservation and movement consistency caused by the insufficient motion representation, we introduce the Pose Indicator, which leverages both implicit and explicit features to enhance the motion understanding of the model. Furthermore, to overcome the limitation of static backgrounds in prior works, we devise a multi-task and partial parameter training strategy that enables text-driven background dynamics. In this way, \method demonstrates strong generalization and robustness, achieving general X character animation in a lively environment.
The proposed framework showcases significant improvements over state-of-the-art methods in terms of identity preservation and motion consistency, as evidenced by experiments on both public datasets and the newly introduced \benchmark, which features anthropomorphic characters.

\section*{Acknowledgements}
This work is supported by the National Natural Science Foundation of China (No. 62422606, 62201484, 624B2124), and Hong Kong Research Grants Council General Research Fund (No. 17213925).

\bibliographystyle{IEEEtran}
\bibliography{main}

\vfill

\end{document}